\documentclass[lettersize,journal]{IEEEtran}
\ifCLASSOPTIONcompsoc
  \usepackage[nocompress]{cite}
\else
  \usepackage{cite}
\fi
\ifCLASSINFOpdf
\else
\fi

\usepackage{bm}
\usepackage{amssymb}
\usepackage{latexsym}
\usepackage{dsfont}
\usepackage{multirow}
\usepackage{amsfonts}
\usepackage{amsmath}
\usepackage{algorithm}
\usepackage{algorithmic}
\usepackage{graphicx}
\usepackage{hyperref}
\usepackage{color}
\usepackage{soul} 

\DeclareMathAlphabet{\mathsfit}{\encodingdefault}{\sfdefault}{m}{sl}
\SetMathAlphabet{\mathsfit}{bold}{\encodingdefault}{\sfdefault}{bx}{n}

\DeclareMathOperator*{\argmax}{arg\,max}

\newcommand{\tabincell}[2]{\begin{tabular}{@{}#1@{}}#2\end{tabular}}

\newcommand{\new}[1]{\textcolor{black}{{#1}}}

\newcommand{\ldz}[1]{\textcolor{black}{{#1}}}

\def\et{{\it et al.}}

\begin{document}
%
\title{Hard-Label Black-Box Attacks \\ on 3D Point Clouds}

\author{Daizong~Liu, Yunbo~Tao, Junhao Dong, Keke Tang,
        Pan~Zhou,~\IEEEmembership{Senior~Member,~IEEE},
        Wei~Hu,~\IEEEmembership{Senior~Member,~IEEE}, and Yew-Soon Ong,~\IEEEmembership{Fellow,~IEEE}
\IEEEcompsocitemizethanks{
\IEEEcompsocthanksitem This work was supported by National Natural Science Foundation of China under Grants  62225113, WHU-Kingsoft Joint Lab, National Natural Science Foundation of China (NSFC) under grant No. 62476107, Shenzhen Science and Technology Program JCYJ20250604191037048, and the New Cornerstone Science Foundation through the XPLORER PRIZE. 
\IEEEcompsocthanksitem Daizong Liu is with the Institute for Math \& AI, Wuhan University, Wuhan, 430072, China. 
E-mail: daizongliu1996@gmail.com.
\IEEEcompsocthanksitem Yunbo Tao and Pan Zhou are with Huazhong University of Science and Technology, Shenzhen Huazhong University of Science and Technology Research Institute, Wuhan 430074, China. E-mail: tyb666@hust.edu.cn, panzhou@hust.edu.cn.
\IEEEcompsocthanksitem Junhao Dong and Yew-Soon Ong are with the College of Computing and Data Science, Nanyang Technological University, Singapore. 
E-mail: junhao003@ntu.edu.sg, asysong@ntu.edu.sg.
\IEEEcompsocthanksitem Keke Tang is with the Cyberspace Institute of Advanced Technology, Guangzhou University, Guangzhou, 510006, China. 
E-mail: tangbohutbh@gmail.com.
\IEEEcompsocthanksitem Wei Hu is with Wangxuan Institute of Computer Technology, Peking University,  Beijing, China. 
E-mail: forhuwei@pku.edu.cn.
\IEEEcompsocthanksitem Corresponding author: Pan Zhou and Wei Hu.}}

%
%

\markboth{Journal of \LaTeX\ Class Files,~Vol.~14, No.~8, August~2015}%
{Shell \MakeLowercase{\textit{et al.}}: Bare Demo of IEEEtran.cls for Computer Society Journals}
%



\IEEEtitleabstractindextext{%
\begin{abstract}
With the maturity of depth sensors in various 3D safety-critical applications, 3D point cloud models have been shown to be vulnerable to adversarial attacks.
Almost all existing 3D attackers simply follow the white-box or black-box setting to iteratively 
update coordinate perturbations based on back-propagated or estimated gradients.
However, these methods are hard to deploy in real-world scenarios (no model details are provided) as they severely rely on parameters or output logits of victim models.
To this end,
we propose point cloud attacks from a more practical setting, \textit{i.e.}, hard-label black-box attack, in which attackers can only access the prediction label of 3D input. 
We introduce a novel 3D attack method based on a new spectrum-aware decision boundary algorithm to generate high-quality adversarial samples.
In particular, we first construct a class-aware model decision boundary, by developing a learnable spectrum-fusion strategy to adaptively fuse point clouds of different classes in the spectral domain, aiming to craft their intermediate samples without distorting the original geometry.
Then, we devise an iterative coordinate-spectrum optimization method with curvature-aware boundary search to move the intermediate sample along the decision boundary for generating adversarial point clouds with trivial perturbations.
Experiments demonstrate that our attack competitively outperforms existing white/black-box attackers in terms of attack performance and adversary quality.
\end{abstract}

}

\maketitle

\IEEEdisplaynontitleabstractindextext

%
\IEEEpeerreviewmaketitle

\section{Introduction}
\IEEEPARstart{D}{eep} Neural Networks (DNNs) are proven to be vulnerable to adversarial examples, which add trivial perturbations to benign samples, making them indistinguishable from legitimate ones but causing the model to produce incorrect prediction.
Significant progress has been made in adversarial attacks on the 2D field, where most methods \cite{madry2017towards,tu2019autozoom} learn to add imperceptible pixel-wise noise in the spatial or feature domain for fooling the 2D models.
Nevertheless, in addition to image-based 2D attacks, adversarial attacks on 3D depth or point-cloud data are still relatively under-explored.
With the maturity of depth sensors, 3D point clouds have received increasing attention in various safety-critical
applications such as autonomous systems \cite{zhang2019eye,zhao2025lossless}, robotic grasping \cite{zhong2020reliable} and biomedical applications \cite{singh20203d}. Similar to their 2D counterparts, these 3D models are also vulnerable to adversarial perturbations, increasing the risk in realistic scenarios. 

Existing 3D point cloud attack methods \cite{tsai2020robust,zhao2020isometry,zhou2020lg,wen2020geometry,xiao2025dual,xie2022stealthy}
typically focus on designing robust attack algorithms with high attack success rates while improving the imperceptibility of adversarial examples.
To generate high-quality adversarial samples, they often adopt geometric distance losses or additional shape knowledge to implicitly constrain point-wise perturbations according to gradient search or gradient optimization.
Although they have achieved significant progress, most of them \cite{xiang2019generating,wen2020geometry,huang2022shape} are deployed in the simple white-box setting where the attackers have the full knowledge of victim models including both network structure and weights. This setting makes the attacks less practical since most real-world 3D applications will not share their model details with users. 
To alleviate this reliance on model details to a certain extent, recent methods \cite{huang2022shape} propose to attack the 3D model under the black-box setting.
However, these works still require the knowledge of the predicted logit scores of the input point cloud to estimate the gradients for back-propagation and optimization.

\begin{figure}[!t]
	\centering
	\includegraphics[width=0.5\textwidth]{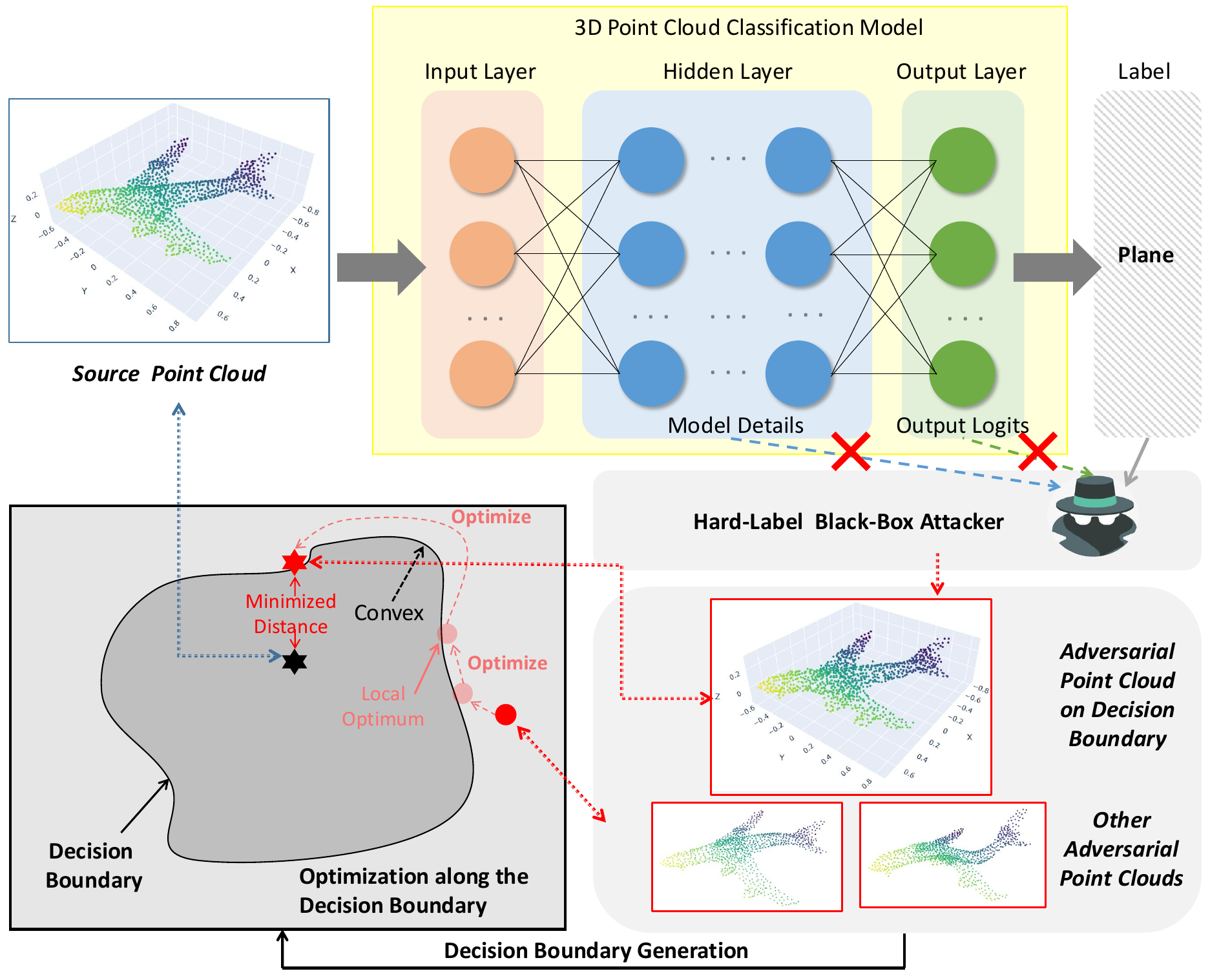}
        \vspace{-10pt}
	\caption{Illustration of our motivation. Our 3D hard-label black-box setting cannot access any model details of the hidden layer or the logits of the output layer. To tackle this setting, we develop a spectrum-aware decision boundary algorithm to first fuse point clouds in the spectral domain for generating the class-aware decision boundary, and then iteratively optimize the best adversarial sample along the decision boundary based on its geometric curvature information and distances.}
        \vspace{-10pt}
	\label{fg:intro}
\end{figure}

To this end, we make the attempt to explore a more practical yet challenging 3D attack setting, \textit{i.e.}, attacking 3D point clouds with {\it black-box hard labels}, in which attackers can only have access to the final classification results of the model without resorting to model details and predicted logits. 
\new{We follow the same targeted attack setting as previous 3D attack methods.}
However, with no prior model knowledge, it is difficult to determine how to guide the adversarial perturbations toward the accurate optimization direction. Fortunately, decision boundary mechanism \cite{li2020qeba,li2022decision,li2021nonlinear} is proven to be effective in handling this hard-label black-box setting, which iteratively queries the classification model to produce the sample-based decision boundary\footnote{In a \new{statistical classification} problem with multiple classes, a decision boundary is a hyper-surface that partitions the underlying vector space into two sets. The classifier will classify all the samples on one side of the decision boundary as belonging to one class and all those on the other side as belonging to the other classes.}, then generates and optimizes adversarial samples based on the characteristic distributions of samples near the decision boundary.
Inspired by this,
we attempt to exploit the decision boundary mechanism as the core idea to generate adversarial point clouds with only the prediction labels, as shown in Figure~\ref{fg:intro}.
\new{While decision-boundary-based attacks\cite{li2022decision,li2020qeba,li2021nonlinear} have achieved decent progress in the 2D image field, to the best of our knowledge, this decision boundary mechanism has been seldom investigated in the 3D adversarial attack, which has the following challenges:
(1) 2D decision boundary attackers \cite{li2020qeba,li2021nonlinear} generally generate adversarial images on the decision boundary by calculating the weighted average of each pixel value between source and target images. However, since points in the 3D space are unordered and irregularly sampled, as for decision boundary sample generation, directly calculating the weighted average of point coordinates between two point clouds would disarrange the relations of neighboring points and deform the 3D object shape.
(2) Previous average-fused decision-boundary images often preserve the semantic features of the original image since their pixel relations are implicitly maintained due to a steganography-like principle \cite{lu2021large,xu2022robust}.
However, the crucial structural representations of 3D point clouds in the latent space would be easily broken when modifying points in the data domain.
(3) Previous 2D decision boundary mechanisms directly utilize pixel-wise iterative walking to move the image along the decision boundary for optimization. However, solely utilizing data-domain constraints to achieve point-wise walking on 3D point clouds may stuck into a local optimum and severely distort the 3D geometries, since the optimized boundary cloud may not have the smallest perturbation and fail to overcome the large convex area without additional guidance.}

In this paper, we propose a novel spectrum-based 3D decision boundary attack method to address the above challenges.
Firstly, instead of fusing the point clouds via naive coordinate-wise average, we propose to merge them in the spectral domain, which is verified to explicitly reveal the geometric structures from basic shapes to fine details via different frequency bands \cite{hu2021graph,hu2022exploring}. By carefully designing the spectral fusion strategy,
we not only preserve the geometric characteristics of both point clouds, but also produce the piecewise-smooth structure of the generated sample in the data domain for achieving high imperceptibility.
Secondly, in addition to the general coordinate-walking during the boundary cloud optimization, we propose to also optimize the boundary cloud in the spectral domain. Since
the boundary cloud may have different coordinate and spectrum
distances from its source/benign cloud, using both walkings in two domains can alleviate the local optimal problem caused by a single domain.

To be specific, our overall method consists of two main stages: 
(1) \emph{Boundary-cloud generation} with learnable spectrum-fusion: we leverage graph spectral tools \cite{hu2021graph,ortega2022introduction} to fuse the point clouds in the frequency domain for boundary cloud generation. We first construct a K-Nearest Neighbor (KNN) graph over the input point clouds and perform Graph Fourier Transforms (GFT) on each point cloud, thus projecting them onto the graph spectral domain with geometry interpretation. 
We then fuse the spectrum of the source (benign) point cloud and target one in terms of the GFT coefficients with learnable fusion rates, and further perform inverse GFTs to generate corresponding candidate point clouds, which serve as coarse boundary clouds.
The learnable fusion rates are dynamically obtained by a pre-trained lightweight model according to the fusion state of different object classes.
In this manner, we obtain the class-aware decision boundary between point clouds of different class labels. 
We then select the best candidate with the slightest distortion and project it on the decision boundary to obtain the initial boundary cloud.
(2) \emph{Boundary-cloud optimization} along the decision boundary: we then perform a joint coordinate and spectrum iterative walking approach on the initial boundary cloud along the decision boundary, aiming at further minimizing its distance to the source cloud for better optimization. 
Each iteration starts from a perturbation generated by gradient estimation and then reduces the distortion through binary searching coordinates and spectra in the adversarial region.
Since the boundary cloud has different coordinate and spectral distances from its benign cloud, this joint walking algorithm is able to alleviate the local optimal problem caused by the concave-convex structure of the decision boundary in a single domain. 
During the above optimization process, we further propose to exploit geometric information of the decision boundary to guide the boundary cloud to move along the curvatures for achieving high imperceptibility.

Our main contributions are summarized as follows:
\begin{itemize}
    \item We introduce a more challenging yet practical \new{hard-label black-box setting} for 3D point cloud attacks, without resorting to any details of victim models. To achieve this, we propose a novel attack paradigm based on learnable class-aware decision boundaries in the spectral domain by iteratively querying and optimizing adversarial point cloud samples. 
    \item To construct the class-aware decision boundary, we propose to fuse point clouds of different classes in the graph spectral domain, aiming to alleviate geometric distortion caused by the straightforward coordinate-wise fusion. In particular, we develop a learnable spectrum-fusion method, which trains and collects desirable fusion weights to generate high-quality initial boundary clouds.
    \item To further optimize the boundary cloud along the decision boundary, we propose iterative walking in both the spatial and spectral domains with further curvature-aware geometric optimization designs, so as to effectively and efficiently search for the optimal cloud with the least and imperceptible perturbation. 
    \item Extensive experiments over widely adopted benchmarks and 3D models demonstrate that, in even the more challenging hard-label setting, \new{our proposed attack method still achieves competitive balance between imperceptibility and efficiency (as in Table~\ref{tab:intro})} compared to existing attack methods in easier white-/black-box settings.
\end{itemize}

\begin{table}[t!]
\caption{\new{Our attack achieves a competitive balance between imperceptibility and efficiency in the challenging hard-label black-box setting (Victim Model: PointNet).}}
\centering
\setlength{\tabcolsep}{1.4mm}{
\begin{tabular}{c|c|ccc}
\hline
\multirow{2}*{\new{Attack}} & \multirow{2}*{\new{Setting}} & \new{Hausdorff} & \new{Chamfer} & \new{Speed Per} \\
~ & ~ & \new{Distance} & \new{Distance} & \new{Sample (s)} \\ \hline
\new{GeoA$^3$ \cite{wen2020geometry}} & \new{White-Box} & \new{0.0175} & \new{0.0064} & \new{1.72} \\ \hline
\new{SI-Adv$^b$\cite{huang2022shape}} & \new{Black-Box} & \new{0.0431} & \new{0.0003} & \new{1.98} \\ \hline
\new{3DHacker \cite{tao20233dhacker}} & \new{Hard-Label} & \new{0.0136} & \new{0.0017} & \new{2.66}\\
\new{Ours} & \new{Black-box} & \new{0.0123} & \new{0.0011} & \new{2.42}\\
\hline
\end{tabular}}
\label{tab:intro}
\end{table}

Some preliminary ideas of this paper have appeared in our earlier work \cite{tao20233dhacker}. In this paper, we extend the previous method from the following three aspects.
Firstly, the previous version manually defined \textit{fixed} fusion rates with handcraft studies for fusing point clouds, resulting in coarse geometric preservation as point-cloud pairs of different classes have different frequency distributions. Instead, our extension designs a \textit{learnable} process to finely and efficiently generate suitable fusion rates for a certain point-cloud pair to construct more natural and imperceptible geometric structures.
Secondly, our previous work directly follows a 2D decision boundary mechanism with normal-vector-based binary search for boundary cloud optimization. However, due to the limited query budget and the non-linearity of the boundary, the estimated normal vector may be inaccurate and result in wrong predictions. Instead, here we propose a curvature-aware optimization approach to adjust the boundary cloud based on the geometric information of the decision boundary, which is more efficient and effective.
Thirdly, we evaluate the proposed attack on more recent victim 3D models for comparison, and add more experimental results to further investigate the effectiveness of our newly proposed components. Compared to our previous work, 
\new{the above extensions show significant advantages in the following specific aspects: (1) Our method achieves significant distortion reductions across various 3D architectures, reaching a 31.8\% improvement on advanced CurveNet and a 17.6\% improvement on advanced SimpleView, which far exceeds the conference version. (2) Moreover, our learnable fusion and {curvature-aware optimization} adaptively scale the perturbation based on local geometry, showing a more specific and promising local structure preservation than the conference version. (3) These numerical gains translate into a qualitative leap in adversarial stealthiness, resulting in a performance gap of up to 21.1\% in ASR when facing advanced geometric defenses—particularly the recent point removal/denoising methods. (4)
\ldz{Our method addresses the critical challenge of {query efficiency} in practical 3D adversarial testing under limited query budgets. This makes it a viable tool for assessing the robustness of commercial 3D APIs and real-time systems where query frequency is strictly monitored or limited.}
Overall, this demonstrates that our extension provides a substantial, non-marginal advancement in both attack potency and manifold consistency.}

\section{Related Work}
\subsection{3D Point Cloud Classification}
3D object classification is a fundamental task that involves extracting representative information from point clouds.
Early attempts at point cloud classification involved firstly mapping the 3D object into multiple 2D views and then utilizing deep-learning models to classify and vote their shapes \cite{yu2018multi}, which entailed max-pooling multi-view object features to create a global representation. However, these methods incurred high computational costs due to the use of numerous 2D convolution layers. To address the challenge of directly extracting 3D structure and mitigating the inherent unordered nature of point clouds, some researchers proposed to transform the input points into a latent space with potentially canonical order, and then apply standard convolutional operations to learn corresponding 3D features.
Diverging from the aforementioned approaches, pioneering 3D classification methods like DeepSets and PointNet \cite{qi2017pointnet} proposed an end-to-end learning paradigm for point cloud classification by formulating a general framework for point cloud analysis. Based on this 3D general backbone with coordinate-aware points input, PointNet++ \cite{qi2017pointnet++} and subsequent works \cite{duan2019structural,liu2019densepoint,yang2019modeling} expanded upon PointNet to capture fine-grained local structural information from each point's neighborhood.
PAConv \cite{xu2021paconv} proposed to incorporate attention mechanisms into the convolutional layers. It can efficiently capture local and global features in the point cloud data.


\subsection{3D Point Cloud Attack}
Following previous studies on the 2D image field, many 3D works \cite{xiang2019generating,wicker2019robustness,zheng2019pointcloud,tsai2020robust,zhao2020isometry,zhou2020lg} adapt adversarial attacks into the 3D vision community \cite{zhao2021point,yu2022point,wang2019dynamic,uy2019revisiting,qi2017pointnet,gao2025distributed,xu2025prediction,fu2026hyperr3snet,chen2025learning,qi2017pointnet++}. 
Most of the existing 3D adversarial attack methods are under the white-box setting, where the attackers have the full knowledge of victim models including network structure and weights.
For example, Xiang \et \cite{xiang2019generating} proposed point generation attacks by adding a limited number of synthesized points/clusters/objects to a point cloud. It adjusted the added points by exploring the gradient vectors according to the white models.
Zhang \et \cite{zhang2019defense} utilized also gradient-guided attack methods to explore more complicated attacks including point modification, addition, and deletion.
Recently, more works \cite{zhang2019defense,tsai2020robust,wen2020geometry,liu2021imperceptible} adopt point-wise perturbation by changing their XYZ coordinates, which are more effective and efficient. These works employ the iterative gradient back-propagation strategy to update the initial perturbations.
Specifically, Liu \et \cite{liu2021imperceptible} modify the FGSM strategy to iteratively search the desired pixel-wise perturbation.
Tsai \et \cite{tsai2020robust} adapted the C\&W strategy to generate adversarial examples on point clouds and proposed a perturbation-constrained regularization in the overall loss function. 
They also deform the mesh-level offsets by modifying the gradient direction.
Besides, some works \cite{kim2021minimal,shi2022shape,dong2022isometric} attack point clouds in the feature space and target at perturbing fewer points with lighter distortions for an imperceptible adversarial attack. 
Although the above white-box attackers have achieved significant attack performance in recent years, these works make the attacks less practical since most real-world 3D applications will not share their model details with users.
To alleviate this limitation, a few works \cite{huang2022shape} proposed to tackle the black-box attack setting, which only accesses the knowledge of predicted logits scores of the input instead of the model details. However, this setting is still not practical since it still relies on the changes of the final predicted logits for updating the perturbations. 

\subsection{Decision Boundary Attack on 2D Image Field}
Decision boundary attack method \cite{brendel2017decision} is widely used in the 2D image field, which is an effective framework that uses the final decision results of a classification model to implement the hard-label black-box attack. To be specific, in the 2D field, the decision boundary attack process starts with two origin images called 
\emph{source-image} and \emph{target-image} with different labels. Then, it exploits a binary search strategy to obtain a \emph{boundary-image} on the decision boundary between \emph{source-image} and \emph{target-image}. Next, a random walking algorithm is conducted on this \emph{boundary-image} to minimize its distance towards \emph{target-image} while maintaining its label the same as \emph{source-image}. 
Based on this general attack framework, various 2D decision boundary attacks are further proposed committed to improve the random walking performance and query efficiency.
In particular,
\cite{brunner2019guessing} propose to choose more efficient random perturbation including Perlin noise and DCT in random walking steps instead of Gaussian perturbation. \cite{chen2020hopskipjumpattack} conduct a gradient estimation method using the Monte-Carlo sampling strategy instead of random perturbation. \cite{li2020qeba,li2021nonlinear,li2022decision} improve the gradient estimation strategy through sampling from representative low-dimensional subspace. 
\cite{li2022decision} first present a basic method to produce candidate adversarial examples in the frequency domain instead of fusing images in the pixel level, then propose a complete boundary attack based on this frequency-mixup method.
However, \new{to the best of our knowledge}, there is no decision boundary based attack been investigated in the 3D vision community so far, which may face many challenges. 

\section{The Proposed 3D Hard-Label Black-Box Attack}
In this section, we elaborate on the proposed hard-label black-box attack for 3D point clouds. 
We first introduce the notations and problem statement of 3D adversarial attacks, and then provide an overview of the proposed decision boundary algorithm to generate high-quality adversarial point clouds without using any model details.
In particular, our attack method involves 1) learnable spectral fusion for generating 3D decision boundary; and 2) the curvature-aware coordinate-spectrum optimization method, which moves the boundary cloud along the decision boundary to the optimal position with trivial perturbations.

\begin{figure*}[t!]
	\centering
	\includegraphics[width=\textwidth]{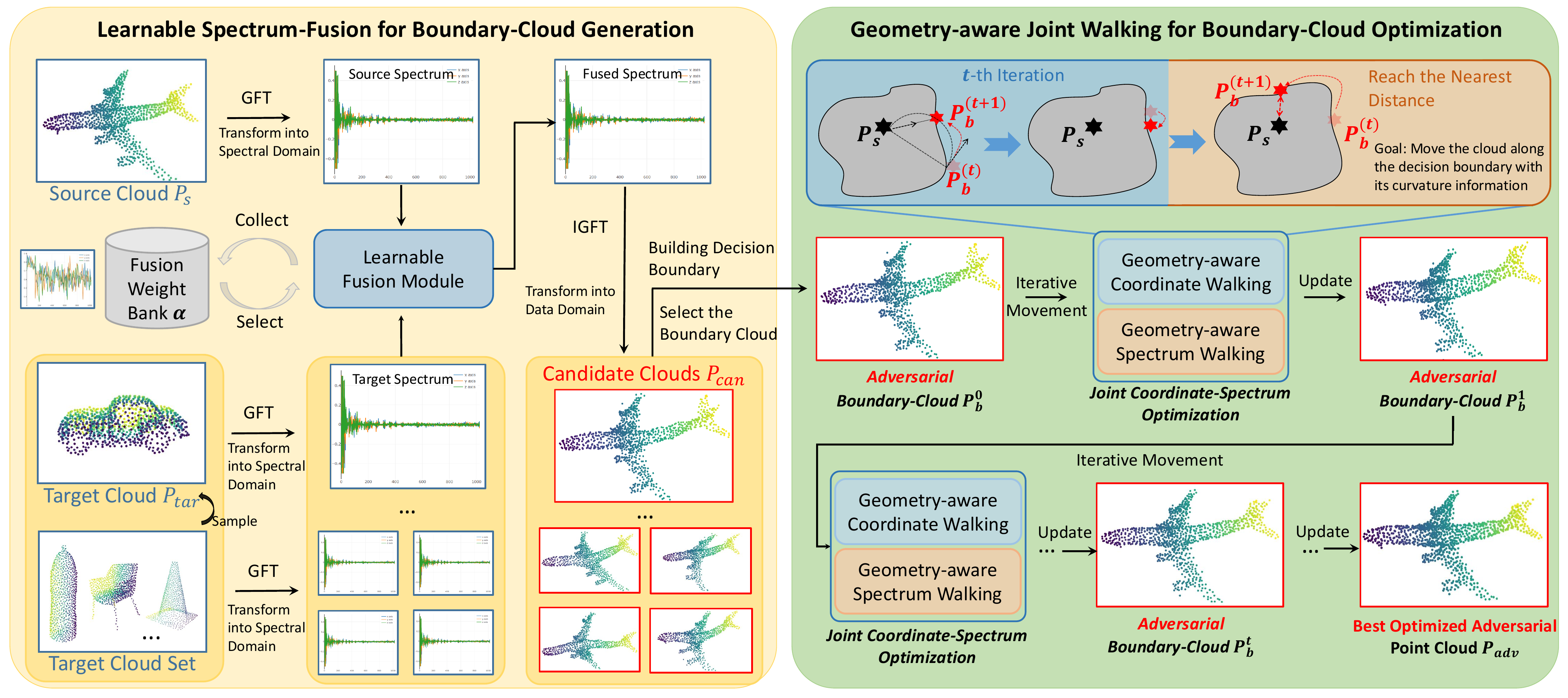}
        \vspace{-10pt}
	\caption{Overall pipeline of our proposed Hard-Label Black-Box Attack. Specifically, we first design a learnable spectrum-fusion boundary-cloud generation module to fuse the source cloud with a set of target clouds in the spectral domain to construct candidate clouds with high \new{imperceptibility}. Then, we utilize the obtained candidate clouds to construct the decision boundary and select the best one to project onto the decision boundary for obtaining the initial boundary cloud. After that, we introduce a geometry-aware boundary-cloud optimization module to iteratively move this boundary cloud along the decision boundary via a joint coordinate-spectrum iterative walking strategy to achieve the best place with smallest distortion. Here, we consider the curvature information of the decision boundary to help update the boundary cloud. The well-optimized boundary cloud is our final generated adversarial point cloud.}
        \vspace{-10pt}
	\label{fig:pipeline}
\end{figure*}

\subsection{Problem Statement}
Given a clean point cloud $\mathbf{P}$ sampled from the surface of a 3D object or scene, we denote its unordered set of points as $\mathbf{P}=\{\mathbf{p}_i\}_{i=1}^n \in \mathbb{R}^{n \times 3}$ where $n$ is the number of points and each point $\mathbf{p}_i \in \mathbb{R}^{3}$ is a vector containing the exact coordinates $(x,y,z)$.
In this paper, we mainly focus on attacking a basic point cloud learning task, \textit{i.e.}, point cloud classification.
In this task, a learned classifier $f(\cdot)$ is provided to predict a vector of confidence scores $f(\mathbf{P})\in {\mathbb{R}^{C}}$ of the input point cloud $\mathbf{P}$, where $C$ is the number of classes.
The final output label is denoted as $y=F(\mathbf{P})=\argmax_{i\in C}f(\mathbf{P})_c\in { Y}, {Y}= \{1,2,3,...,C\}$, which represents a certain class of the original 3D object underlying the point cloud.

To attack this 3D classifier, a general objective \cite{xiang2019generating,huang2022shape} is to find a perturbation $\mathbf{\Delta} \in \mathbb{R}^{n \times3}$ to generate a corresponding adversarial example as $\mathbf{P}_{adv} = \mathbf{P} + \mathbf{\Delta}$. These works follow white/black-box settings to use the model parameters $f(\cdot)$ or the confidence scores (logits) $f(\mathbf{P})$ to optimize the gradients of the network for updating $\mathbf{\Delta}$. 
However, in our focused hard-label black-box attack setting, we only have access to the final predicted label $y=F(\mathbf{P})$ without using any model detail.
To this end, we aim to query the 3D models with multiple constructed point clouds to build the class-aware decision boundary \new{(\textit{i.e.}, pairwise decision boundary constructed specifically between a certain source class and other target classes)} for determining the difference between different-class 3D objects.
Based on the generated decision boundary of a specific object class, we initialize a boundary point cloud and estimate the positive gradient to update its added perturbation by iteratively moving it along the decision boundary.
That is, the perturbed adversarial sample $\mathbf{P}_{adv}$ is carefully generated and optimized from the correctly classified source sample $\mathbf{P}_s$ labeled as $y_{gt}$ along the decision boundary, such that \new{the model output attacker-chosen label $F(\mathbf{P}_{adv})=y_{adv}\ne y_{gt}$}. \new{We adopt a targeted attack setting in our cases as it is more challenging than the untargeted attack setting \cite{liu2021imperceptible,liu2022point}}. Here, we define an indicator function $\varphi(\cdot)$ of a successful attack as:
\new{\begin{equation}\label{eq1}
\varphi(\mathbf{P}_{adv})\equiv\begin{cases}
1,  & \text{if } F(\mathbf{P}_{adv})=y_{adv},  \\
-1, & \text{if } F(\mathbf{P}_{adv})\ne y_{adv},
\end{cases}
\end{equation}}
where 
$y_{adv}$ is the \new{attacker-chosen} label of the corresponding adversarial point cloud $\mathbf{P}_{adv}$.

\subsection{Overview of Our Decision Boundary Approach}
We introduce a novel attack method to tackle the challenging 3D hard-label attack setting. Since attackers cannot access any model detail, we propose to generate a boundary cloud on the model decision boundary between the source and target point clouds as the adversarial sample, which has a class label different from the source cloud while sharing similar object shape.
The overall pipeline is illustrated in Figure~\ref{fig:pipeline}. 
Here, our hard-label black-box attack method mainly consists of two stages: 
1) a \textbf{boundary-cloud generation} module is first proposed to produce a high-quality adversarial point cloud on the decision boundary; and 
2) a \textbf{boundary-cloud optimization} module is exploited to further optimize the adversarial point cloud along the decision boundary, aiming to achieve smallest perturbations. 

\subsubsection{First Stage: Boundary-Cloud Generation}
\noindent \textbf{Principle and challenges.} 
\new{A boundary sample/cloud is constructed on the decision boundary, which is ambiguous to classifiers.
In the 2D attack field, previous 2D decision boundary mechanisms generally fuse the source and target images via pixel-wise average operation to generate boundary samples. 
Since neighboring pixels are often smooth and the images share the same size, this operation would not degrade the image quality significantly and the fused sample tends to remain the semantics of the original images.
However, 3D point cloud data mainly contain point-wise coordinates that are irregularly sampled and unordered. 
Hence, the pixel-wise operation is inapplicable to the 3D domain, since 3D coordinate-wise fusion between point clouds will result in the outlier problem and destroy the 3D object geometric shape, leading to poor imperceptibility.}

\noindent \textbf{The key idea.}
To address the above issue, we introduce a novel learnable spectrum-fusion strategy, which fuses point clouds in the spectral domain so as to preserve the geometric topology for improving imperceptibility. Instead of fusing point-wise coordinates, our spectrum-fusion method leverages graph spectral tools \cite{hu2021graph,ortega2022introduction} to first transform both source and target point clouds into the spectral domain for representing their geometric characteristics, then adaptively fuses their spectral contexts according to their classes in a learnable way and transforms the fused geometric characteristics back to the data domain as the generated adversarial sample. 
In particular, since the spectral coefficients of a point cloud can generally be divided into two bands, we separately fuse the low- and high-frequency components of two point clouds in the spectral domain to preserve the basic 3D object shape (low-frequency) and fine-grained details (high-frequency). 
In this manner, the fused point cloud not only preserves specific characteristics of original point clouds, but also has smoother geometric surface due to spectral fusion \cite{hu2021graph,hu2022exploring}.

\subsubsection{Second Stage: Boundary-Cloud Optimization}
\noindent \textbf{Principle and challenges.}
\new{After obtaining the boundary sample on the decision boundary in the first stage, previous 2D decision boundary mechanisms often optimize the boundary sample along the decision boundary to search for a high-quality position that has the lowest geometric distance to the source sample. Specifically, they generally conduct the iterative walking strategy with pixel-wise offsets to adjust the images. However, this data-domain walking cannot be directly applied to the 3D attack domain since point-wise coordinate walking will lead to local optimum object shape due to the specific concave-convex structure of the decision boundary.
The reason is, although the boundary cloud can be well optimized and still remains adversarial, it may get stuck in the local concave area with a large neighboring convex area, and the estimated gradients towards this convex hull will lead to a larger geometric distance to the source point cloud.} 
Besides, previous 2D decision boundary mechanisms solely utilize a simple binary search strategy based on normal vectors for boundary cloud optimization. However, due to the limited query budget and the non-linearity of the boundary, the estimated normal vector may be inaccurate and result in wrong predictions. Moreover, this optimization process is time-consuming.

\noindent \textbf{The key idea.} 
To alleviate the above issues, we propose to extend the point-wise coordinate walking with a novel spectrum walking as additional guidance for jumping out of such local optimum. 
We argue that the decision boundary reflected in the spectral domain is different from that in the data domain, since the boundary cloud likely exhibits different distances to the original cloud in the spatial and spectral domain.
Therefore, we update the boundary cloud along the decision boundary with both coordinate modification for shape refinement and spectral modification for maintaining geometric smoothness and high imperceptibility.
Further, to increase the query efficiency, we propose a curvature-aware optimization strategy to adjust the boundary cloud along a semicircular path inspired by \cite{maho2021surfree}.
In this way, the boundary cloud can be better optimized to achieve high quality and imperceptibility.

In the following, we will elaborate on the process of each stage.

\subsection{Learnable Spectrum-Fusion for Boundary-Cloud Generation}
To generate the boundary cloud on the decision boundary, we propose to fuse two point clouds in the spectral domain. 
For the source clouds of a specific class label, we assume that they share the same class-sensitive tolerance to the noise when we add the target clouds to them. Therefore, we design a learnable spectrum-fusion strategy to first train and collect desired fusion weights of a certain class via a lightweight learnable model with further adversarial learning. 
Then, during the inference, we directly employ the collected fusion weights to generate high-quality boundary clouds by fusing the source and target clouds.

\begin{figure}[!t]
	\centering
	\includegraphics[width=0.49\textwidth]{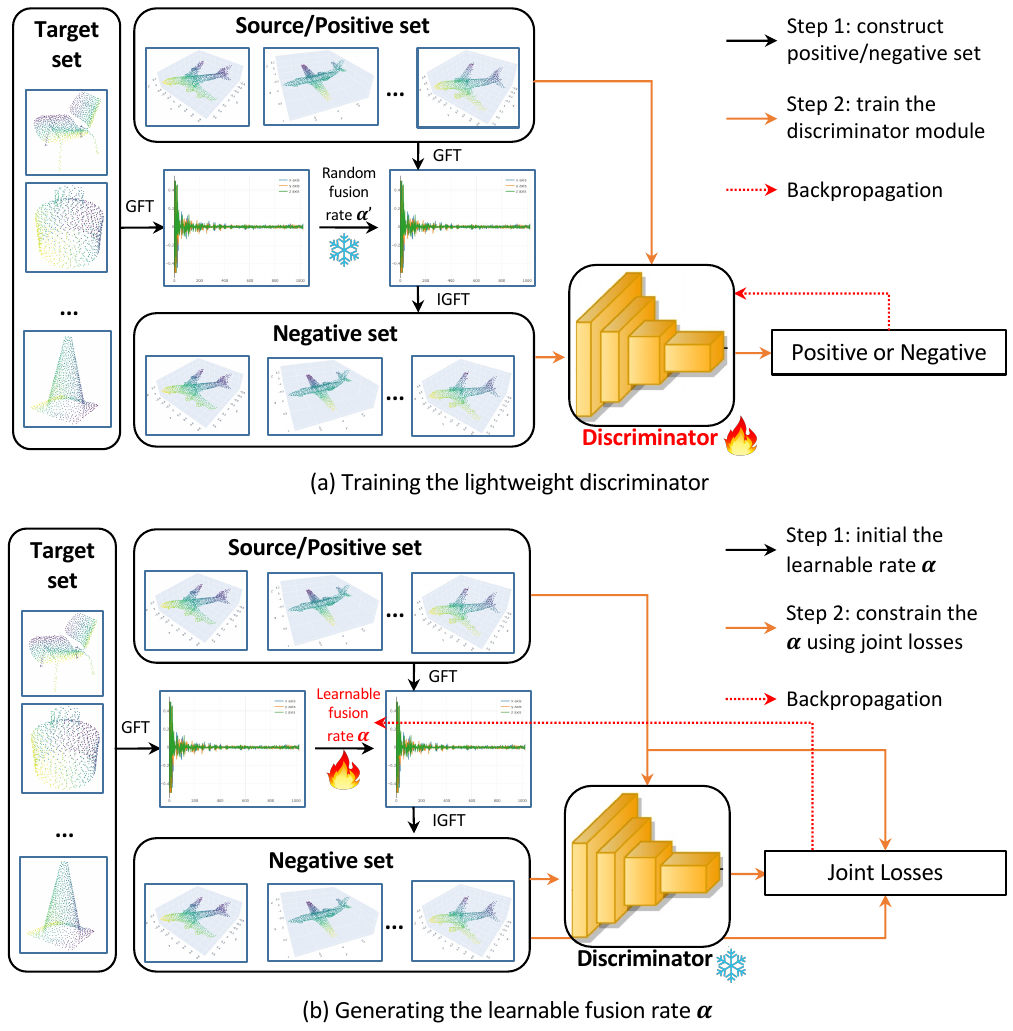}
        \vspace{-10pt}
	\caption{Illustration of how we generate the learnable spectrum-fusion rates. We first design a discriminator to distinguish the benign and randomly fused point clouds. Then, we initial a learnable rate $\mathbf{\alpha}$ and utilize the gradients of discriminator to backpropagate and update it.}
        \vspace{-10pt}
	\label{fig:spectrum}
\end{figure}

\noindent \textbf{Learning fusion weights $\mathbf{\alpha}$.}
To obtain the optimal fusion weights without manual tuning, it is crucial to train a model to backpropagate the gradients so as to guide the weight update.
However, in our setting, we cannot utilize any details of existing 3D models.
Therefore, we propose a self-learning strategy to develop a lightweight discriminator module, which distinguishes the distributions of fused point clouds from their original ones. Specifically, given the source point clouds $\{\mathbf{P}_s\}$ of a specific class and target point clouds $\{\mathbf{P}_{tar}\}$ of different classes, we take their fused samples into a negative set and the source point clouds into a positive set.
In particular, for each pair of $\mathbf{P}_s$ and $\mathbf{P}_{tar}$, we follow \cite{hu2021graph} to conduct Graph Fourier Transform (GFT) to obtain their corresponding spectral coefficient vector $\hat{\mathbf{x}}_s$ and ${\hat{\mathbf{x}}_{tar}}$ for spectral fusion:
\begin{equation}
\begin{aligned}
\hat{\mathbf{x}}_{low} &=\mathbf{\alpha}_{low} {GFT}(\mathbf{P}_s)_L+(1-\mathbf{\alpha}_{low}){GFT}(\mathbf{P}_{tar})_L, \\
\hat{\mathbf{x}}_{high}&=\mathbf{\alpha}_{high} {GFT}(\mathbf{P}_s)_H+(1-\mathbf{\alpha}_{high}){GFT}(\mathbf{P}_{tar})_H,
\end{aligned}
\end{equation}
where $GFT(\mathbf{P}_s)_L$ and $GFT(\mathbf{P}_s)_H$ denote the low- and high-frequency coefficients of $\mathbf{P}_s$, respectively. 
The same goes to $GFT(\mathbf{P}_{tar})_L$ and $GFT(\mathbf{P}_{tar})_H$. 
$\mathbf{\alpha}_{low}$ and $\mathbf{\alpha}_{high}$ are the \textit{random} fusion weights.
We piece two coefficient vectors $\hat{\mathbf{x}}_{low},\hat{\mathbf{x}}_{high}$ into a complete one and
perform inverse GFT (IGFT) to transform the fused spectral coefficient vector back to the data domain.
After constructing both the positive and negative sets, we utilize the two-sample classifier to train the discriminator module $\mathbb{I}(\cdot)$ similar to the principle of the generative adversarial networks. We use the simple point cloud encoder to encode each point cloud into a latent vector and employ three linear layers to predict whether it is positive or negative. The detailed process is shown in Figure~\ref{fig:spectrum} (a).

After obtaining the trained discriminator, for each class of the point clouds, we employ an adversarial learning strategy to learn their optimal fusion weights with the trained discriminator. Their learned fusion weights are subsequently collected into a weight bank, representing the class-specific frequency tolerance to the noise. 
In this manner, we directly fetch one of the fusion weights for specific class fusion during the inference. 
To achieve this adversarial learning, as shown in Figure~\ref{fig:spectrum} (b), we denote the learnable fusion weights as $\mathbf{\alpha}$ and define the loss function as:
\begin{equation}
\begin{aligned}
    & \min \mathcal{L}_{class}(\mathbf{P}_s') + \mathcal{L}_{dis}(\mathbf{P}_s,\mathbf{P}_s') + \mathcal{L}_{reg}(\mathbf{P}_s,\mathbf{P}_s'), \\
    & \text{s.t.} \ \ \mathbf{P}_s' = IGFT(\mathbf{\alpha} GFT(\mathbf{P}_s)+(1-\mathbf{\alpha})GFT(\mathbf{P}_{tar})).
\end{aligned}
\end{equation}
Here, $\mathcal{L}_{class}$ guides the discriminator $\mathbb{I}(\cdot)$ to predict positive result on the fused sample $\mathbf{P}_s'$; $\mathcal{L}_{dis}$ can be any geometric distance to restrict the geometric shape of $\mathbf{P}_s'$. 
Considering that directly deploying the Chamfer distance \cite{fan2017point} or the Hausdorff distance \cite{huttenlocher1993comparing} is too strict and may limit the diversity of the class shape, we follow the 1-nearest neighbor accuracy (1-NNA) strategy \cite{yang2019pointflow} to achieve a soft distance constraint as:
\begin{equation}
    \mathcal{L}_{dis}(\mathbf{P}_s') = - \frac{\sum_{I \in \{\mathbf{P}_s'\}}\mathbb{I}(N_I)}{|\{\mathbf{P}_s'\}|},
\end{equation}
where $|\{\mathbf{P}_s'\}|$ denotes the number of fused negative sample set $\{\mathbf{P}_s'\}$, $N_I$ is the nearest neighbor of each negative sample among both positive and negative sets.
$\mathcal{L}_{reg}(\mathbf{P}_s,\mathbf{P}_s')$ is a low-frequency constraint \cite{liu2022point} to limit perturbations within imperceptible details, which guides the perturbation to concentrate on the high-frequency components that represent fine details and noise.
The final learned weight $\mathbf{\alpha}$ will be collected in the weight bank.

\noindent \textbf{Generating boundary-clouds with the collected weights.}
During the inference, we directly randomly choose the collected weights in the weight bank to fuse the source point clouds $\mathbf{P}_s$ of a specific class and $T$ number of target point clouds $\{\mathbf{P}_{tar}\}$.
\textit{Although we can generate new fusion weights for each source-target cloud pair, this process costs additional time and we find that its performance is only slightly better than using the weight bank via experiments.}
We denote their $T$ fused samples as candidate clouds $\{\mathbf{P}_{can}\}$, and conduct the adversarial evaluation on them to select the best one for further optimization as the final boundary cloud $\mathbf{P}_b$. 
To be specific, we first eliminate non-adversarial candidate clouds where $\varphi(\mathbf{P}_{can}^l)=-1$ and select the best candidate cloud $\mathbf{P}_{can}^B$ from the adversarial candidate clouds with the slightest distortion measured by distance metric $D(\mathbf{P}_s,\mathbf{P}_{can})$ as:
\begin{equation}
\label{eq:epsilon}
\begin{aligned}
&\mathbf{P}_{can}^B = \text{argmin}_{\mathbf{P}_{can}^l} D(\mathbf{P}_s,\mathbf{P}_{can}^l)(l=1,2,...,T), \quad\\
&\text{s.t.} \ \varphi(\mathbf{P}_{can}^l)=1,
\text{max}_j(\left\| \mathbf{p}_{can,i}^l-\mathbf{p}_{s,i}\right\|_2)\le\varepsilon,
\end{aligned}
\end{equation}
where $\mathbf{P}_{can}^l$ denotes the $l$-th candidate-cloud, $\mathbf{p}_{can,i}^l$ denotes the $i$-th points in $\mathbf{P}_{can}^l$, $\mathbf{p}_{s,i}\in\mathbf{P}_s$ denotes the point with the lowest distance from $\mathbf{p}_{can,i}^l$. $\varepsilon$ is utilized to select cloud samples without local outliers.
The global distance measurement function $D(\cdot)$ is formulated as:
\begin{equation}
\label{eq:gamma}
\begin{aligned}
D(\mathbf{P}_s,&\mathbf{P}_{can}^l)= D_{Chamfer}(\mathbf{P}_s,\mathbf{P}_{can}^l) \\ & +\hfill 
\gamma_1D_{Hausdorff}(\mathbf{P}_s,\mathbf{P}_{can}^l)  +\gamma_2D_{L2Norm}(\mathbf{P}_s,\mathbf{P}_{can}^l),
\end{aligned}
\end{equation}
where $D_{Chamfer}$ and $D_{Hausdorff}$ measure the distance between two point clouds following \cite{wen2020geometry}. $D_{L2Norm}$ is the L2-Norm distance between two point clouds. $\gamma_1$ and $\gamma_2$ are penalty parameters. 
After obtaining the optimal candidate cloud $\mathbf{P}_{can}^B$, we project it onto the model decision boundary to obtain the boundary-cloud $\mathbf{P}_{b}$ via binary search strategy by iteratively moving $\mathbf{P}_{can}^B$ towards $\mathbf{P}_{s}$ until reaching the decision boundary as:
\begin{equation}
\begin{split}
&\mathbf{P}_b^h=\beta\mathbf{P}_s+(1-\beta)\mathbf{P}_{can}^{B},\\
&\text{until} \ \varphi(\mathbf{P}_b^{h})=1 \ \text{and} \ \varphi(\mathbf{P}_b^{h+1})=-1,
\end{split}
\end{equation}
where $\beta$ is the moving ratio, $h$ denotes the iteration step.

\begin{figure}[!t]
	\centering
	\includegraphics[width=0.49\textwidth]{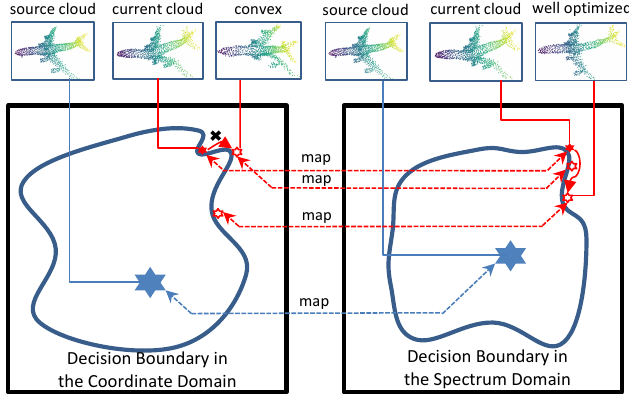}
        \vspace{-10pt}
	\caption{Illustration of why we propose to optimize the boundary cloud in both coordinate and spectrum domains. The boundary cloud may have different coordinate and spectrum distances from the original cloud. Therefore, the point cloud in the convex position of the coordinate domain may no longer be a convex position in the spectrum domain and can be well-optimized. In this manner, we can avoid the local optimal problem in the coordinate domain by additionally optimizing the cloud in the spectrum domain.}
        \vspace{-10pt}
	\label{fig:boundary}
\end{figure}

\subsection{Geometry-aware Coordinate-Spectrum Walking for Boundary-Cloud Optimization}
To further optimize the boundary cloud obtained in the first stage along the decision boundary, we propose an iterative walking algorithm in both coordinate and spectral domains to search for the optimal place so that the perturbation is the smallest yet imperceptible. Since the boundary cloud has different coordinate and spectrum distances from its benign cloud, as shown in Figure~\ref{fig:boundary}, this joint walking algorithm is able to alleviate the local optimal problem caused by the concave-convex structure of the decision boundary in a certain domain. During the above optimization process, instead of using complex normal vectors based binary search walking, we further propose to exploit geometric information of the decision boundary to guide the boundary cloud to move along the curvatures.

\noindent \textbf{Joint coordinate-spectrum walking.}
The goal of boundary-cloud optimization is to minimize the distance between the boundary cloud $\mathbf{P}_b$ and the source cloud $\mathbf{P}_s$ by moving $\mathbf{P}_b$ along the decision boundary to the optimal place. 
To achieve this goal, we design the iterative walking algorithm in both coordinate and spectrum domains to optimize the point cloud $\mathbf{P}_b$.

Specifically, we first take the previously obtained $\mathbf{P}_b$ as the initial boundary cloud $\mathbf{P}_b^{(0)}$. Then, for the next step, we denote $\mathbf{P}_b^{(t)}$ as the boundary cloud obtained in the $t$-th walking iteration, which is exactly on the decision boundary. We aim to update $\mathbf{P}_b^{(t)}$ to improve the gap between the adversarial and the true class labels while preserving their geometric distance, so that we can make $\mathbf{P}_b^{(t)}$ more aggressive with small distortion and can further move it towards source cloud $\mathbf{P}_s$. 
In particular, we first employ the \textit{coordinate walking} to move $\mathbf{P}_b^{(t)}$ in the data domain by:
\begin{equation}
\mathbf{P}_b^{(t+1)}=\phi(\mathbf{P}_s, \mathbf{P}_b^{(t)}),
\end{equation}
where $\phi$ denotes the walking strategy. Generally, $\phi$ is defined as the Monte Carlo method \cite{james1980monte} that estimates and utilizes the gradient vector for optimizing $\mathbf{P}_b^{(t)}$ with normal vector based binary search strategy. 

To alleviate the local optimum problem, in addition to the \textit{coordinate walking strategy}, we also conduct a \textit{spectrum walking} to bring a great movement for cloud features from the aspect of the spectral domain so as to escape the convex area in the data domain.
Specifically, we solely replace the point-wise operation with a frequency-aware operation and maintain other key-point walking operations:
\begin{equation}
\mathbf{P}_b^{(t+1)}=IGFT(\phi(GFT(\mathbf{P}_s), GFT(\mathbf{P}_b^{(t)}))).
\end{equation}
By combining two walking operations for multi-step walking, the spectrum walking is able to perform large movement to jump out of local optima for a better optimization region, while the coordinate walking is able to perform slight movement to gradually fine-tune for acquring the best optimization point in such region. 
The optimized point cloud is our final adversarial sample.

\begin{figure}[!t]
	\centering
	\includegraphics[width=0.42\textwidth]{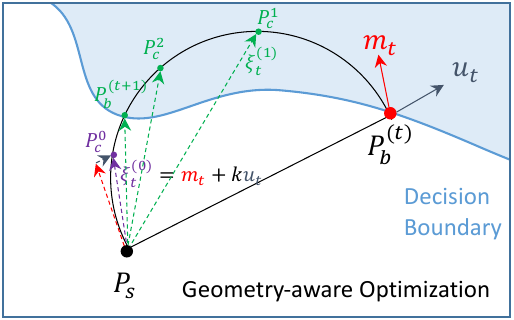}
        \vspace{-4pt}
	\caption{Illustration on our geometry-aware optimization strategy. Specifically, given the cloud $\mathbf{P}_b^{(t)}$, we iteratively update and adjust the direction $\mathbf{\xi}_t$ to search the cloud $\mathbf{P}_c$ along the semicircular path until reaching the best adversarial position $\mathbf{P}_b^{(t+1)}$.}
        \vspace{-10pt}
	\label{fig:curvature}
\end{figure}

\noindent \textbf{Geometry-aware optimization.}
In the above joint coordinate-spectrum walking process, a simple way to achieve the walking strategy is to utilize the Monte Carlo method with binary search to define $\phi$ for boundary cloud optimization. However, this strategy relies on a large number of cloud queries and is not efficient. Therefore, we propose to update each step cloud $\mathbf{P}_b^{(t)}$ solely based on the geometric curvatures of the decision boundary.

As shown in Figure~\ref{fig:curvature}, given the source cloud $\mathbf{P}_s$ and $t$-th step boundary cloud $\mathbf{P}_b^{(t)}$, we propose to conduct a boundary search to obtain a better subsequent boundary cloud $\mathbf{P}_b^{(t+1)}$ on a semicircular path \cite{maho2021surfree}, where $\mathbf{P}_b^{(t+1)}$ has smaller distance to $\mathbf{P}_s$. Note that, different from \cite{maho2021surfree}, we employ additional normal vector information for better guidance. Moreover, our semicircular path and corresponding center are formed between $\mathbf{P}_s$ and $\mathbf{P}_b^{(t)}$, which can be employed in both coordinate and spectral domains. Specifically, we denote $\mathbf{u}_t$ as the direction of $\mathbf{P}_b^{(t)}$ from $\mathbf{P}_s$. $\mathbf{m}_t$ is the estimated normal direction on $\mathbf{P}_b^{(t)}$, which is calculated by:
\begin{equation}
\label{eq:m}
\begin{aligned}
     &\text{Coordinate:} \ \mathbf{m}_t =\frac{\sum_{i=1}^B \varphi(\mathbf{P}_b^{(t)}+\mathbf{v}_i)\mathbf{v}_i}{||\sum_{i=1}^B \varphi(\mathbf{P}_b^{(t)}+\mathbf{v}_i)\mathbf{v}_i||_2}, \\
     &\text{Spectral:} \ \mathbf{m}_t = \frac{\sum_{i=1}^B \varphi (IGFT(GFT(\mathbf{P}_b^{(t)})+\mathbf{v}_i))\mathbf{v}_i}{||\sum_{i=1}^B \varphi (IGFT(GFT(\mathbf{P}_b^{(t)})+\mathbf{v}_i))\mathbf{v}_i||_2},
\end{aligned}
\end{equation}
where $\mathbf{v}_i$ is the sampled move vector obeying a normal distribution and $B$ is the corresponding number.
Therefore, we obtain the initial search direction $\mathbf{\xi}_t^{(0)}$ to perform a query for the non-adversarial point cloud $\mathbf{P}_c^{(0)}$ on the semicircle near the decision boundary in the plane spanned by $(\mathbf{u}_t,\mathbf{m}_t)$ by:
\begin{equation}
    \mathbf{\xi}_t^{(0)} = \frac{k\mathbf{u}_t + \mathbf{m}_t}{||k\mathbf{u}_t + \mathbf{m}_t||_2},
\end{equation}
where $k=\frac{1}{2^q}, \forall{q\in\mathbb{N}}$ is a search parameter inspired by the binary search. With the increase of $q$, the search direction $\mathbf{\xi}_t^{(0)}$ is getting closer to $\mathbf{m}_t$ until the searched point cloud $\mathbf{P}_c^{(0)}$ is non-adversarial. This $\mathbf{P}_c^{(0)}$ can be obtained by:
\begin{equation}
    \mathbf{P}_c^{(0)} = \mathbf{P}_s+ {||\mathbf{P}_b^{(t)} - \mathbf{P}_s||_2}{(\mathbf{\xi}_t^{(0)} \cdot \mathbf{u}_t)}\mathbf{\xi}_t^{(0)}.
\end{equation}

\begin{algorithm}[t!]
\caption{Geometry-aware Joint Coordinate-Spectrum Walking} \label{alg1}
{\bf Input:} Boundary cloud $\mathbf{P}_b^{(0)}$, iteration step $R$, best optimized cloud list $Best=[]$. \\
{\bf Output:} Adversarial cloud $\mathbf{P}_{adv}$.
\begin{algorithmic}[1] 
\STATE $Best[0]=\mathbf{P}_b^{(0)}$;
\STATE {\bf for} $t=1:R$ {\bf do}
\STATE \qquad {\bf while} True {\bf do} \quad \#conduct coordinate walking
\STATE \qquad \qquad initialize $\mathbf{\xi}_t^{(j-1)}$ in data domain;
\STATE \qquad \qquad update $\mathbf{\xi}_t^{(j)}$ with $\mathbf{P}_c^{(j)}$;
\STATE \qquad \qquad {\bf if} $\varphi(\mathbf{P}_c^{(j)})+\varphi(\mathbf{P}_c^{(j+1)})=0$ {\bf then}
\STATE \qquad \qquad \qquad $\mathbf{P}_b^{(t+1)}=\mathbf{P}_c^{(j)}$;
\STATE \qquad \qquad \qquad {\bf break}
\STATE \qquad {\bf if} $\mathbf{P}_b^{(t)}=\mathbf{P}_b^{(t+1)}$ {\bf then}
\STATE \qquad \qquad {\bf while} True {\bf do} \quad \#conduct spectrum walking
\STATE \qquad \qquad \qquad initialize $\mathbf{\xi}_t^{(j-1)}$ in spectral domain;
\STATE \qquad \qquad \qquad update $\mathbf{\xi}_t^{(j)}$ with $\mathbf{P}_c^{(j)}$;
\STATE \qquad \qquad \qquad {\bf if} $\varphi(\mathbf{P}_c^{(j)})+\varphi(\mathbf{P}_c^{(j+1)})=0$ {\bf then}
\STATE \qquad \qquad \qquad \qquad $\mathbf{P}_b^{(t+1)}=\mathbf{P}_c^{(j)}$;
\STATE \qquad \qquad \qquad \qquad {\bf break}
\STATE \qquad {\bf else}
\STATE \qquad \qquad $Best\text{.append}(\mathbf{P}_b^{(t+1)})$\;   
\STATE {\bf end} 
\STATE Select $\mathbf{P}_{adv} \in Best$ with the smallest distance;
\end{algorithmic}
\end{algorithm}

After finding the non-adversarial $\mathbf{P}_c^{(0)}$, we then progressively optimize the adversarial cloud along the semicircle to obtain the best boundary cloud $\mathbf{P}_b^{(t+1)}$ by adjusting $\mathbf{\xi}_t$. 
Taking Figure~\ref{fig:curvature} as an example, we initialize a lower bound direction $\mathbf{\xi}_{lower}=\mathbf{\xi}_t^{(0)}$ and an upper bound direction $\mathbf{\xi}_{upper}=\mathbf{u}_t$. Then during the $j$-step optimization, we obtain the search direction $\mathbf{\xi}_t^{(j)}=\frac{\mathbf{\xi}_{lower}+\mathbf{\xi}_{upper}}{||\mathbf{\xi}_{lower}+\mathbf{\xi}_{upper}||_2}$ and the cloud $\mathbf{P}_c^{(j)}$. If $\varphi(\mathbf{P}_c^{(j)})=1$, we adjust $\mathbf{\xi}_{upper}=\mathbf{\xi}_t^{(j)}$, else we adjust $\mathbf{\xi}_{lower}=\mathbf{\xi}_t^{(j)}$.
This process of reducing the range of the
search direction is continued until obtaining the boundary
cloud $\mathbf{P}_b^{(t+1)}$ with a certain accuracy. One important characteristic of this process is that it ensures $\mathbf{P}_b^{(t+1)}$ with a reduced perturbation for any query on the semicircular path.
Since this geometry-aware decision boundary optimization does not rely on a large number of queries of the previous strategy \cite{tao20233dhacker}, it is much more efficient by solely querying on the geometric curvature of the decision boundary. The overall geometry-aware coordinate-spectrum walking algorithm is detailed in Algorithm \ref{alg1}.

\begin{table*}[t!]
\caption{Comparative results on the perturbation sizes of adversarial point clouds generated by different attack methods under 100\% ASR.}
\centering
\setlength{\tabcolsep}{1.4mm}{
\begin{tabular}{c|c|cc|c|ccc|ccc|ccc}
\hline
\multirow{2}*{Setting} & \multirow{2}*{Attack} & \multicolumn{2}{c|}{Model Details} & \multirow{2}*{ASR(\%)} & \multicolumn{3}{|c}{PointNet \cite{qi2017pointnet}} & \multicolumn{3}{|c}{PointNet++ \cite{qi2017pointnet++}} & \multicolumn{3}{|c}{DGCNN \cite{wang2019dynamic}}\\
\cline{3-4} \cline{6-14}
~ & ~ & Para. & Logits &~ &$D_h$ & $D_c$ & $D_{norm}$ &$D_h$& $D_c$ & $D_{norm}$ & $D_h$ & $D_c$ & $D_{norm}$ \\
\hline
\multirow{10}*{White-Box}
&FGSM \cite{yang2019adversarial}&$\checkmark$&$\checkmark$&100&0.1853&0.1326&0.7936 &0.2275&0.1682&0.8357 &0.2506&0.1890&0.8549 \\
&PGD \cite{madry2017towards}&$\checkmark$&$\checkmark$&100&0.1322&0.1329&0.7384&0.1623&0.1138&0.7596&0.1546&0.1421&0.7756 \\
&AdvPC \cite{hamdi2020advpc}&$\checkmark$&$\checkmark$&100&0.0343&0.0697&0.6509&0.0429&0.0685&0.6750&0.0148&0.0623&0.6304\\
&3D-ADV$^p$ \cite{xiang2019generating}&$\checkmark$&$\checkmark$&100&0.0105&0.0003&0.5506 &0.0381&0.0005&0.5699&0.0475&0.0005&0.5767\\
&LG-GAN \cite{zhou2020lg}&$\checkmark$&$\checkmark$&100&0.0362&0.0347&0.7184&0.0407&0.0238&0.6896&0.0348&0.0119&0.8527\\
&GeoA$^3$ \cite{wen2020geometry}&$\checkmark$&$\checkmark$&100&0.0175&0.0064&0.6621&0.0357&0.0198&0.6909&0.0402&0.0176&0.7024\\
&SI-Adv$^w$ \cite{huang2022shape}&$\checkmark$&$\checkmark$&100&0.0204&0.0002&0.7614&0.0310&0.0004&1.2830&0.0127&0.0006&1.1120\\
&\new{HIT-Adv \cite{lou2024hide}} &\new{$\checkmark$}&\new{$\checkmark$}&\new{100}&\new{0.0271}&\new{0.0039}&\new{0.8519} &\new{0.0321}&\new{0.0052}&\new{0.9124} &\new{0.0245}&\new{0.0048}&\new{0.8876}\\
&\new{NoPain \cite{li2025nopain}}&\new{$\checkmark$}&\new{$\checkmark$}&\new{100}&\new{0.0152} &\new{0.0022} &\new{0.7015} &\new{0.0284} &\new{0.0035} &\new{0.8241} &\new{0.0189} &\new{0.0028} &\new{0.7356}\\
&\new{CoSA \cite{tang2026rethinking}} &\new{$\checkmark$}&\new{$\checkmark$}&\new{100}&\new{0.0107}&\new{0.0017}&\new{0.6816} &\new{0.0255}&\new{0.0026}&\new{0.7892} &\new{0.0114}&\new{0.0019}&\new{0.7022}\\
\hline
Black-Box&SI-Adv$^b$ \cite{huang2022shape}&$\times$&$\checkmark$&100&0.0431&0.0003&0.9351&0.0444&0.0003&1.0857&0.0336&0.0004&0.9081\\
\hline
\multirow{2}*{\tabincell{c}{Hard-Label\\ Black-Box}}
&\multirow{1}*{3DHacker \cite{tao20233dhacker}}&\multirow{1}*{$\times$}&\multirow{1}*{$\times$}&\multirow{1}*{100}&\multirow{1}*{0.0136}&\multirow{1}*{0.0017}&\multirow{1}*{0.8561}&\multirow{1}*{0.0245}&\multirow{1}*{0.0023}&\multirow{1}*{0.9324}&\multirow{1}*{0.0129}&\multirow{1}*{0.0026}&\multirow{1}*{0.9030}\\
~&\multirow{1}*{Ours}&\multirow{1}*{$\times$}&\multirow{1}*{$\times$}&\multirow{1}*{100}&\multirow{1}*{\textbf{0.0123}}&\multirow{1}*{\textbf{0.0011}}&\multirow{1}*{\textbf{0.8112}}&\multirow{1}*{\textbf{0.0209}}&\multirow{1}*{\textbf{0.0015}}&\multirow{1}*{\textbf{0.8871}}&\multirow{1}*{\textbf{0.0125}}&\multirow{1}*{\textbf{0.0024}}&\multirow{1}*{\textbf{0.7568}}\\
\hline \hline
\multirow{2}*{Setting} & \multirow{2}*{Attack} & \multicolumn{2}{c|}{Model Details}&\multirow{2}*{ASR(\%)} & \multicolumn{3}{|c}{PAConv \cite{xu2021paconv}} & \multicolumn{3}{|c}{SimpleView \cite{goyal2021revisiting}} & \multicolumn{3}{|c}{CurveNet \cite{xiang2021walk}}\\
\cline{3-4} \cline{6-14}
~ & ~ & Para. & Logits&~ &$D_h$ & $D_c$ & $D_{norm}$ &$D_h$& $D_c$ & $D_{norm}$ & $D_h$ & $D_c$ & $D_{norm}$ \\
\hline
\multirow{10}*{White-Box}
&FGSM \cite{yang2019adversarial}&$\checkmark$&$\checkmark$&100&0.1769&0.1318&0.7852&0.2014&0.1688&0.8247&0.2291&0.1806&0.8325\\
&PGD \cite{madry2017towards}&$\checkmark$&$\checkmark$&100&0.1246&0.1159&0.7231&0.1428&0.1235&0.6980&0.1307&0.1316&0.7432 \\
&AdvPC \cite{hamdi2020advpc}&$\checkmark$&$\checkmark$&100&0.0276&0.0623&0.6431&0.0395&0.0697&0.6428&0.0259&0.0647&0.6530\\
&3D-ADV$^p$ \cite{xiang2019generating}&$\checkmark$&$\checkmark$&100&0.0039&0.0003&0.5071&0.0348&0.0007&0.5325&0.0446&0.0005&0.5612\\
&LG-GAN \cite{zhou2020lg}&$\checkmark$&$\checkmark$&100&0.0358&0.0314&0.7022&0.0367&0.0201&0.6639&0.0315&0.0107&0.8143\\
&GeoA$^3$ \cite{wen2020geometry}&$\checkmark$&$\checkmark$&100& 0.0159&0.0058&0.6341&0.0285&0.0173&0.7172&0.0396&0.0184&0.6833\\
&SI-Adv$^w$ \cite{huang2022shape}&$\checkmark$&$\checkmark$&100&0.0097&0.0004&0.6920&0.0256&0.0014&2.1522&0.0199&0.0006&0.9803\\
&\new{HIT-Adv \cite{lou2024hide}} &\new{$\checkmark$}&\new{$\checkmark$}&\new{100} &\new{0.0142}&\new{0.0042}&\new{0.8415} &\new{0.0289}&\new{0.0062}&\new{0.9541} &\new{0.0233}&\new{0.0051}&\new{0.9214}\\
&\new{NoPain \cite{li2025nopain}}&\new{$\checkmark$}&\new{$\checkmark$}&\new{100} &\new{0.0084}&\new{0.0018}&\new{0.7236} &\new{0.0212}&\new{0.0028}&\new{0.8152} &\new{0.0155}&\new{0.0022}&\new{0.8041}\\
&\new{CoSA \cite{tang2026rethinking}} &\new{$\checkmark$}&\new{$\checkmark$}&\new{100} &\new{0.0052}&\new{0.0012}&\new{0.6582} &\new{0.0185}&\new{0.0021}&\new{0.7514} &\new{0.0122}&\new{0.0018}&\new{0.7135}\\
\hline
Black-Box&SI-Adv$^b$ \cite{huang2022shape}&$\times$&$\checkmark$&100&0.0449&0.0004&1.3386&0.0469&0.0010&1.8754&0.0453&0.0004&1.4336\\
\hline
\multirow{2}*{\tabincell{c}{Hard-Label\\ Black-Box}}
&\multirow{1}*{3DHacker \cite{tao20233dhacker}}&\multirow{1}*{$\times$}&\multirow{1}*{$\times$}&\multirow{1}*{100}&\multirow{1}*{0.0046}&\multirow{1}*{0.0014}&\multirow{1}*{0.9444}&\multirow{1}*{0.0136}&\multirow{1}*{0.0029}&\multirow{1}*{1.6150}&\multirow{1}*{0.0125}&\multirow{1}*{0.0022}&\multirow{1}*{1.2332}\\
~&\multirow{1}*{Ours}&\multirow{1}*{$\times$}&\multirow{1}*{$\times$}&\multirow{1}*{100}&\multirow{1}*{\textbf{0.0037}}&\multirow{1}*{\textbf{0.0012}}&\multirow{1}*{\textbf{0.8579}}&\multirow{1}*{\textbf{0.0112}}&\multirow{1}*{\textbf{0.0021}}&\multirow{1}*{\textbf{1.4286}}&\multirow{1}*{\textbf{0.0103}}&\multirow{1}*{\textbf{0.0015}}&\multirow{1}*{\textbf{1.2171}}\\
\hline
\end{tabular}}
\vspace{-10pt}
\label{tab:comparison}
\end{table*}

\section{Experiments}

\subsection{Dataset and 3D Models}
\noindent  \textbf{Dataset.} Following previous works, we use ModelNet40 in our experiments to evaluate the attack performance. Specifically, the ModelNet40 dataset consists of 12,311 CAD models from 40 object categories, in which 9,843 models are intended for training and the other 2,468 for testing. Following the settings of previous works \cite{wen2020geometry,liu2021imperceptible,hu2022exploring}, we also uniformly sample 1024 points from the surface of each object and scale them into a unit ball. For the adversarial point cloud attacks, we follow \cite{xiang2019generating,hu2022exploring} to achieve fair comparison and randomly select 25 instances for each of 10 object categories in the ModelNet40 testing set, which can be well classified by the classifiers of interest.

\noindent  \textbf{3D Models.} We use six 3D networks in the 3D computer vision community as the victim models: PointNet \cite{qi2017pointnet}, PointNet++ \cite{qi2017pointnet++}, DGCNN \cite{wang2019dynamic}, PAConv \cite{xu2021paconv}, SimpleView \cite{goyal2021revisiting}, and CurveNet \cite{xiang2021walk}. We train them from scratch, and the test accuracy of each trained model is within 0.1\% of the best accuracy reported in their original articles.

\subsection{Implementation Details}
\noindent \textbf{Attack Setup.}
\new{We follow previous works to utilize the same targeted attack setting for fair comparison, as it is more challenging than the untargeted one.}
As for point cloud spectrum-fusion, we set K = 10 to build a K-NN graph to conduct GFTs.
To train the discriminator, for each class,  we train for 100 epochs with a learning rate of 0.002 using the Adam optimizer. 
As for collecting the weight bank, we train each fusion rate for 50 epochs with a learning rate of 0.001.
While generating the adversarial examples, the weights of Hausdorff distance loss and L2-Norm loss, \textit{i.e.}, $\gamma_1$ and $\gamma_2$ in Eq.~\ref{eq:gamma} are set to 2.0 and 0.5, respectively. The outlier limitation $\epsilon$ of Eq.~\ref{eq:epsilon}. is set to 0.2.
We use \textit{B}= $30\sqrt{t}$ queries to obtain the estimated normal direction $\mathbf{m}_t$ in Eq.~\ref{eq:m}. We conduct $R=100$ iteration rounds during boundary-cloud optimization stage. 

\begin{figure}[!t]
	\centering
	\includegraphics[width=0.42\textwidth]{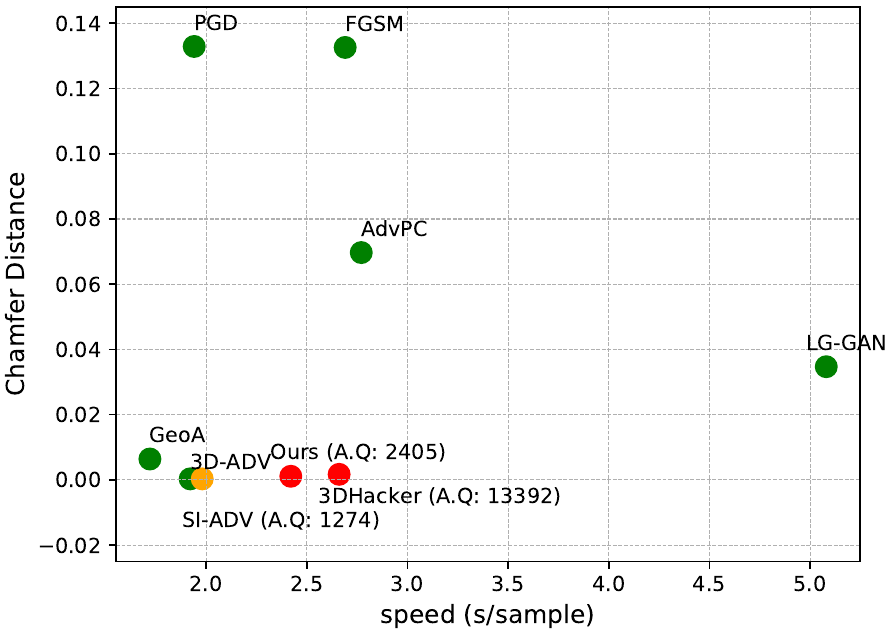}
        \vspace{-14pt}
	\caption{Comparisons on inference speed and the perturbation size. Here, the ``speed" denotes the average time for each adversarial point cloud generation\new{, ``A.Q" denotes average queries}. Green nodes: White-box attackers; Orange nodes: Black-box attackers; Red nodes: Hard-label black-box attackers.}
        \vspace{-10pt}
	\label{fig:efficiency}
\end{figure}

\noindent \textbf{Evaluation Metrics.}
To quantitatively evaluate the effectiveness of our proposed attack, we measure the perturbation size via three metrics: (1) L2-norm distance $D_{norm}$, which measures the squared root of the sum of squared shifting distance; 
(2) Chamfer distance $D_c$ \cite{fan2017point}, which measures the average squared distance between each adversarial point and its nearest original point; 
(3) Hausdorff distance $D_h$ \cite{huttenlocher1993comparing}, which measures the maximum squared distance between each adversarial point and its nearest original point and is thus sensitive to outliers.

\subsection{Evaluation on Our Hard-Label Attack}

\begin{figure*}[!t]
	\centering
	\includegraphics[width=1.0\textwidth]{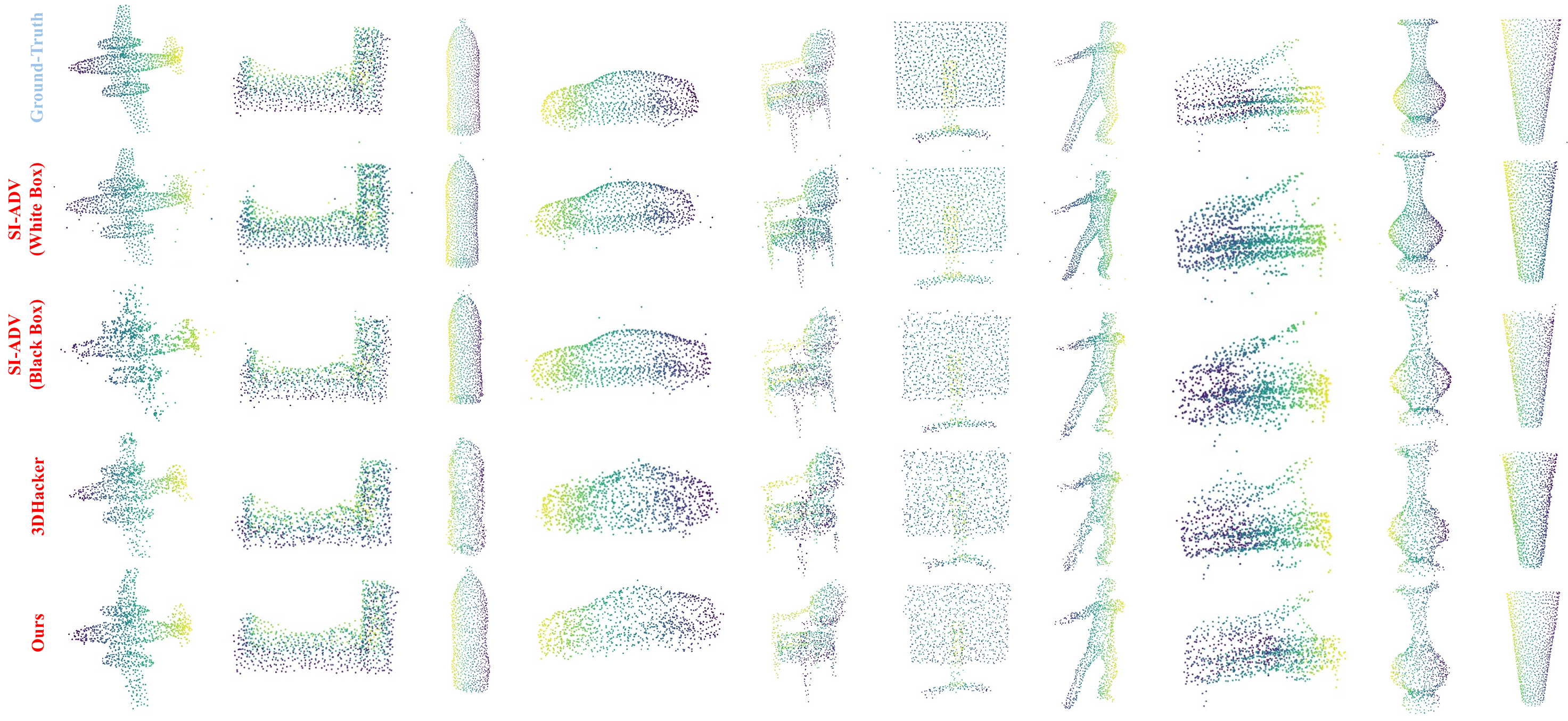}
        \vspace{-14pt}
	\caption{Visualization results of adversarial samples generated by different attack methods. Here, we compare our attack with the SOTA attack methods SI-Adv \cite{huang2022shape} and 3DHacker \cite{tao20233dhacker}. We can find that our samples keep better geometric shapes with less outliers.}
	\label{fig:result}
\end{figure*}

\begin{figure*}[!t]
	\centering
	\includegraphics[width=1.0\textwidth]{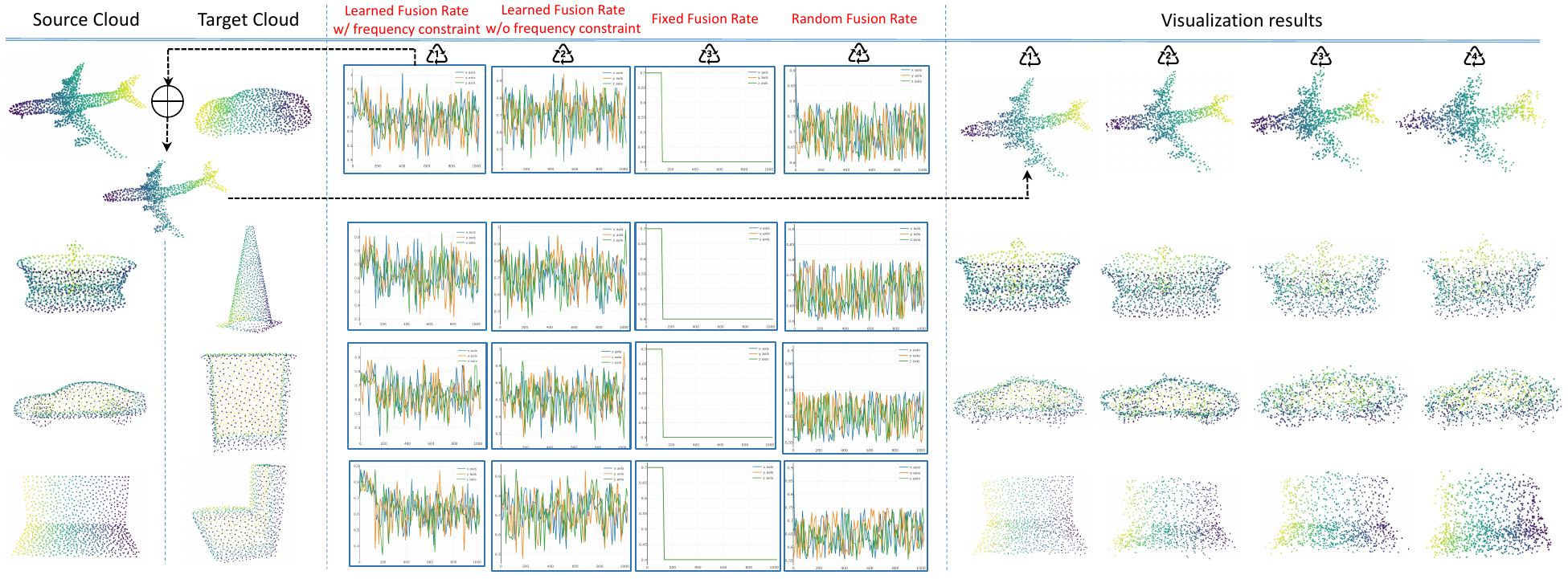}
        \vspace{-10pt}
	\caption{Visualization results of adversarial samples generated by different fusion methods. Here, we list four fusion variants: (1) Learnable fusion with low-frequency constraint; (2) Learnable fusion without fusion constraint; (3) Fixed fusion; (4) Random fusion.}
        \vspace{-10pt}
	\label{fig:fusion}
\end{figure*}

\begin{table*}[t!]
\caption{Resistance of the black-box attacks on defended point cloud models.}
\label{tab:defense}
\centering
\setlength{\tabcolsep}{1.4mm}{
\begin{tabular}{c|c|cccc|cccc|cccc}
\hline
\multirow{2}*{Defense} & \multirow{2}*{Attack} &\multicolumn{4}{|c}{PointNet \cite{qi2017pointnet}} & \multicolumn{4}{|c}{PointNet++ \cite{qi2017pointnet++}} & \multicolumn{4}{|c}{DGCNN \cite{wang2019dynamic}}\\
\cline{3-14}
~ & ~ & ASR(\%) & $D_h$ & $D_c$& $D_{norm}$  & ASR(\%) & $D_h$ & $D_c$ & $D_{norm}$ &ASR(\%) & $D_h$ &$D_c$& $D_{norm}$ \\
\hline
\multirow{3}*{\new{UPP \cite{ai2025upp}}} & \new{SI-Adv$^b$ \cite{huang2022shape}} & \new{70.4} & \new{0.0462} & \new{0.0014} & \new{2.8105} & \new{61.2} & \new{0.0489} & \new{0.0019} & \new{2.8431} & \new{58.3} & \new{0.0441} & \new{\textbf{0.0018}} & \new{1.6924} \\ 
~ & \new{3DHacker \cite{tao20233dhacker}}& \new{80.2} & \new{0.0152} & \new{0.0019} & \new{1.4210} & \new{72.5} & \new{0.0284} & \new{0.0032} & \new{1.4022} & \new{68.9} & \new{0.0225} & \new{0.0035} & \new{1.5204} \\ 
~ & \new{Ours} & \new{\textbf{88.1}} & \new{\textbf{0.0135}} & \new{\textbf{0.0012}} & \new{\textbf{1.0924}} & \new{\textbf{81.4}} & \new{\textbf{0.0261}} & \new{\textbf{0.0018}} & \new{\textbf{1.0855}} & \new{\textbf{78.2}} & \new{\textbf{0.0194}} & \new{{0.0026}} & \new{\textbf{1.1837}} \\
\hline
\multirow{3}*{\new{N2S3D \cite{li2025learning}}} & \new{SI-Adv$^b$ \cite{huang2022shape}} & \new{65.8} & \new{0.0435} & \new{\textbf{0.0013}} & \new{2.9014} & \new{55.4} & \new{0.0452} & \new{\textbf{0.0022}} & \new{1.5230} & \new{50.7} & \new{0.0385} & \new{\textbf{0.0012}} & \new{1.7542} \\ 
~ & \new{3DHacker \cite{tao20233dhacker}}& \new{78.4} & \new{0.0112} & \new{0.0025} & \new{1.3250} & \new{70.1} & \new{0.0235} & \new{0.0045} & \new{1.4128} & \new{75.6} & \new{0.0142} & \new{0.0034} & \new{1.3025} \\ 
~ & \new{Ours} & \new{\textbf{91.2}} & \new{\textbf{0.0104}} & \new{{0.0015}} & \new{\textbf{0.9542}} & \new{\textbf{84.5}} & \new{\textbf{0.0205}} & \new{{0.0025}} & \new{\textbf{1.0931}} & \new{\textbf{85.4}} & \new{\textbf{0.0131}} & \new{{0.0028}} & \new{\textbf{1.0824}} \\
\hline
\multirow{3}*{LPC \cite{li2022robust}} & SI-Adv$^b$ \cite{huang2022shape} & 82.1 & 0.0458 & 0.0012 & 2.7804 & 72.4 & 0.0473 & \textbf{0.0014} & 2.7635 & 65.3 & 0.0421 & \textbf{0.0016} & 1.5804 \\ 
~ & 3DHacker \cite{tao20233dhacker}& 84.5 & 0.0146 & 0.0018 & 1.3519 & 75.8 & 0.0270 & 0.0029 & 1.3782 & 71.8 & 0.0213 & 0.0031 & 1.4652\\ 
~ & Ours &\textbf{89.2} & \textbf{0.0131} & \textbf{0.0011} & \textbf{1.0857} & \textbf{82.3} & \textbf{0.0254}&0.0016&\textbf{1.0542}&\textbf{79.6}&\textbf{0.0188}&0.0024&\textbf{1.1359}\\
\hline
\multirow{3}*{SOR\cite{zhou2019dup}} & SI-Adv$^b$ \cite{huang2022shape} &89.7 & 0.0420  & \textbf{0.0009}& 3.0193
& 78.9 &0.0436 & 0.0025& 1.3843 
& 72.0 & 0.0341 &\textbf{0.0009}& 1.6480\\ 
~ & 3DHacker \cite{tao20233dhacker}&90.4 & \textbf{0.0100} &0.0023& 1.2486
& 82.7& 0.0218 &0.0043& 1.3759
& 85.4 & 0.0124 &0.0031& 1.2387\\ 
~ & Ours&\textbf{93.8} & 0.0109 & 0.0013 & \textbf{0.9262} & \textbf{86.0} & \textbf{0.0213} & \textbf{0.0021} & \textbf{1.0644}&\textbf{88.3}&\textbf{0.0120}&0.0025&\textbf{1.0536}\\
\hline
\multirow{3}*{Drop(30\%)} & SI-Adv$^b$ \cite{huang2022shape} &96.9 & 0.0426 &\textbf{0.0003}& 1.3680 
& 70.1 & 0.0473 &0.0023& 1.4538 
& 71.2 & 0.0400 &\textbf{0.0004}& 0.8598\\  
~ & 3DHacker \cite{tao20233dhacker}& 97.2 & 0.0179 &0.0016& 0.8391
& 71.3 & 0.0298 &0.0031& 1.2810
& 78.5 & 0.0195 &0.0033& 1.1742\\ 
~ & Ours &\textbf{99.1}&\textbf{0.0138}&0.0012&\textbf{0.8237}&\textbf{76.5}&\textbf{0.0247}&\textbf{0.0020}&\textbf{1.0236}&\textbf{85.7}&\textbf{0.0163}&0.0025&\textbf{0.7942}\\
\hline
\multirow{3}*{Drop(50\%)} & SI-Adv$^b$ \cite{huang2022shape} &93.6 & 0.0420 &\textbf{0.0002}&  1.3844 
& 67.6 &0.0501 &\textbf{0.0013}& 1.9193 
& 75.2 & 0.0358 &\textbf{0.0004}& \textbf{0.6992}\\  
~ & 3DHacker \cite{tao20233dhacker}&95.4 & 0.0182 &0.0023& 0.8328
& 77.4 &0.0285 &0.0032& 1.4735
& 76.8 & 0.0172 &0.0036& 1.2914\\ 
~ & Ours & \textbf{97.3} & \textbf{0.0146}&0.0015&\textbf{0.8194}&\textbf{82.8}&\textbf{0.0239}&0.0016&\textbf{1.1348}&\textbf{84.5}&\textbf{0.1437}&0.0026&0.8231\\
\hline
\hline
\multirow{2}*{Defense} & \multirow{2}*{Attack} &\multicolumn{4}{|c}{PAConv \cite{xu2021paconv}} & \multicolumn{4}{|c}{SimpleView \cite{goyal2021revisiting}} & \multicolumn{4}{|c}{CurveNet \cite{xiang2021walk}}\\
\cline{3-14}
~ & ~ & ASR(\%) & $D_h$ & $D_c$ & $D_{norm}$ & ASR(\%) & $D_h$ & $D_c$ & $D_{norm}$ &ASR(\%) & $D_h$ & $D_c$ & $D_{norm}$ \\
\hline
\multirow{3}*{\new{UPP \cite{ai2025upp}}} & \new{SI-Adv$^b$ \cite{huang2022shape}} & \new{72.5} & \new{0.0485} & \new{0.0018} & \new{2.7104} & \new{70.2} & \new{0.0512} & \new{0.0024} & \new{3.1205} & \new{68.5} & \new{0.0452} & \new{0.0016} & \new{2.8231} \\ 
~ & \new{3DHacker \cite{tao20233dhacker}}& \new{85.4} & \new{0.0145} & \new{0.0019} & \new{1.4502} & \new{82.3} & \new{0.0185} & \new{0.0035} & \new{1.9542} & \new{81.7} & \new{0.0168} & \new{0.0042} & \new{1.5432} \\ 
~ & \new{Ours} & \new{\textbf{92.5}} & \new{\textbf{0.0068}} & \new{\textbf{0.0015}} & \new{\textbf{1.0531}} & \new{\textbf{91.2}} & \new{\textbf{0.0142}} & \new{\textbf{0.0021}} & \new{\textbf{1.5242}} & \new{\textbf{87.4}} & \new{\textbf{0.0135}} & \new{\textbf{0.0015}} & \new{\textbf{1.2854}} \\
\hline
\multirow{3}*{\new{N2S3D \cite{li2025learning}}} & \new{SI-Adv$^b$ \cite{huang2022shape}} & \new{68.4} & \new{0.0384} & \new{\textbf{0.0006}} & \new{2.1524} & \new{62.5} & \new{0.0402} & \new{\textbf{0.0012}} & \new{3.2452} & \new{60.4} & \new{0.0384} & \new{\textbf{0.0009}} & \new{2.6843} \\ 
~ & \new{3DHacker \cite{tao20233dhacker}}& \new{82.1} & \new{0.0035} & \new{0.0014} & \new{0.7542} & \new{80.4} & \new{0.0102} & \new{0.0042} & \new{1.2531} & \new{78.5} & \new{\textbf{0.0112}} & \new{0.0031} & \new{1.3250} \\ 
~ & \new{Ours} & \new{\textbf{94.6}} & \new{\textbf{0.0029}} & \new{{0.0013}} & \new{\textbf{0.6841}} & \new{\textbf{92.1}} & \new{\textbf{0.0094}} & \new{{0.0028}} & \new{\textbf{1.1205}} & \new{\textbf{89.2}} & \new{{0.0125}} & \new{{0.0018}} & \new{\textbf{1.2140}} \\
\hline
\multirow{3}*{LPC \cite{li2022robust}} & SI-Adv$^b$ \cite{huang2022shape} & 85.3 & 0.0478 & 0.0015 & 2.6436 &82.4 &0.0490&0.0021&2.9834&80.2&0.0439&0.0014&2.7028 \\ 
~ & 3DHacker \cite{tao20233dhacker}&87.9 & 0.0132&0.0017&1.3695&88.1&0.0167&0.0031&1.8915&84.5&0.0151&0.0036&1.4257\\ 
~ & Ours & \textbf{93.1} & \textbf{0.0063}&\textbf{0.0014}&\textbf{1.0240}&\textbf{93.5}&\textbf{0.0136}&\textbf{0.0019}&\textbf{1.4893}&\textbf{89.6}&\textbf{0.0125}&\textbf{0.0013}&\textbf{1.2392} \\ 
\hline
\multirow{3}*{SOR \cite{zhou2019dup}} & SI-Adv$^b$ \cite{huang2022shape} 
&94.4 & 0.0359 & \textbf{0.0005} & 1.9640
& 95.2 &0.0375 & \textbf{0.0009} & 3.1333 
& 88.8 & 0.0351 & \textbf{0.0007} & 2.5402\\ 
~ & 3DHacker \cite{tao20233dhacker}&95.5 & 0.0028 & 0.0011 & 0.6744 
& 93.6 & \textbf{0.0083} & 0.0034 & 1.0873 
& 89.2 & \textbf{0.0095} & 0.0023 & 1.1752\\ 
~ & Ours &\textbf{97.7}&\textbf{0.0027}&0.0012&\textbf{0.6549}&\textbf{97.3}&0.0089&0.0024&\textbf{1.0837}&\textbf{94.0}&0.0114&0.0015&\textbf{1.1563} \\
\hline
\multirow{3}*{Drop(30\%)} & SI-Adv$^b$ \cite{huang2022shape} 
&73.6 & 0.0402 & \textbf{0.0005} &  1.1979
& 56.8 & 0.0411 & \textbf{0.0010} &  1.2577
& 71.2 & 0.0400 & \textbf{0.0008} & 1.4630\\  
~ & 3DHacker \cite{tao20233dhacker}& 95.2 & 0.0061 & 0.0017 & 0.8290
& 91.2 & 0.0092 & 0.0042 &  0.9638
& 82.5 & 0.0157 & 0.0034 & 0.8598\\ 
~ & Ours &\textbf{96.4}&\textbf{0.0043}&0.0015&\textbf{0.7964}&\textbf{94.9}&\textbf{0.0083}&0.0025&\textbf{0.9594}&\textbf{88.3}&\textbf{0.0126}&\textbf{0.0008}&\textbf{0.8469} \\
\hline
\multirow{3}*{Drop(50\%)} & SI-Adv$^b$ \cite{huang2022shape} 
&84.8 & 0.0390 & \textbf{0.0004} &   0.8537
& 68.8 & 0.0368 & \textbf{0.0006} & 0.9119
& 79.2 & 0.0392 & \textbf{0.0007} & 1.1759\\  
~ & 3DHacker \cite{tao20233dhacker}& 93.8 & 0.0136 & 0.0022 & \textbf{0.7261}
& 97.6 & \textbf{0.0066} & 0.0037 & 0.7570 
& 83.4 & 0.0186 & 0.0031 & \textbf{0.7558}\\ 
~ & Ours&\textbf{96.2}&\textbf{0.0069}&0.0017&0.7970&\textbf{98.2}&0.0070&0.0023&\textbf{0.7459}&\textbf{90.5}&\textbf{0.0157}&0.0016&0.8322\\
\hline
\end{tabular}}
\end{table*}

\noindent \textbf{Quantitative Comparison.}
To investigate the effectiveness of our proposed hard-label black-box attack, we compare our method with \new{ten} existing white-box adversarial attacks (including FGSM \cite{yang2019adversarial}, PGD \cite{madry2017towards}, AdvPC \cite{hamdi2020advpc}, 3D-ADV$^p$ \cite{xiang2019generating}, LG-GAN \cite{zhou2020lg}, GeoA$^3$ \cite{wen2020geometry}, SI-Adv$^w$ \cite{huang2022shape}\new{, HIT-Adv \cite{lou2024hide}, NoPain \cite{li2025nopain}, CoSA \cite{tang2026rethinking}}) and one black-box adversarial attack (SI-Adv$^b$ \cite{huang2022shape}) for quantitative comparison, where we measure their perturbation in the data domain with three evaluation metrics when these methods reach 100\% of attack success rate (ASR). 
We also report our previous conference version \cite{tao20233dhacker} for comparison.
We implement the above attacks on six 3D models, and the corresponding results are presented in Table~\ref{tab:comparison}. 
From this table, we see that our proposed attack achieves smaller perturbation sizes than the black-box model and achieves very competitive results with white-box models.
Note that, our hard-label black-box setting is much harder to achieve success since it has no information on model details (white box) and output logits (black box).
Since our attack method conducts global perturbations to original point clouds which possess a strong potential to confuse the victim models with a structure distortion, these global perturbations produce a higher Chamfer distance $D_c$ compared with 3D-ADV$^p$ and SI-Adv, because $D_c$ measures the average squared distance between each adversarial point and its nearest original point and we modify all the points leading to a large sum of displacements. Instead, attacking by modifying a few points in 3D-ADV$^p$ has the advantage in $D_c$ because most of the distance is equal to 0. However, our method performs better in $D_h$ since we conduct relatively average perturbations to the point cloud, which does not count on a few outliers to confuse the victim models, leading to imperceptibility and having the potential to bypass the outlier detection defense.
Moreover, compared to our previous conference version 3DHacker \cite{tao20233dhacker}, our new version introduces a more effective learnable fusion module and a better geometry-aware optimization strategy, which improves the quality and imperceptibility of the adversarial example, leading to lower perturbation sizes.

\noindent \textbf{Efficiency Comparison.}
Solely conducting the quantitative comparison to evaluate the effectiveness is not enough, since efficiency is also important in measuring the practicability of attack methods in real-world scenarios.
Therefore, we also provide the running-time experiments to evaluate the attack efficiency of different attack methods. 
As shown in Figure~\ref{fig:efficiency}, our running time is very competitive to the black-box model since our optimization steps can also be efficiently achieved. \new{Also, our required average query number is also competitive compared to black-box attacks.} Some white-box models are the most time-consuming since they need complicated back propagation through the victim model. GeoA and 3D-ADV achieve the fastest speed as their optimization strategies are simple. Compared to our previous version 3DHacker, our new attack method is more efficient as we design a geometry-aware optimization strategy instead of using the time-consuming normal-vector-based binary search strategy.

\ldz{\textit{Discussion:} The query budget is a pivotal metric for evaluating the real-world threat of 3D adversarial attacks, as it directly determines the feasibility of an attack in restricted environments. Our focus on minimizing queries (2405 vs. 13392) is grounded in several critical industry-relevant scenarios. {First}, commercial 3D cloud services—such as the Google Cloud 3D Data API \cite{googlestreetview} and Sketchfab\cite{spiess2024sketchfab}—provide widespread 3D recognition and processing capabilities. These platforms typically implement rate-limiting or anomaly detection mechanisms to block brute-force adversarial probing. Consequently, a low-query attack like ours is essential for performing {stealthy} and {cost-effective} security auditing without triggering defensive alarms. {Second}, in safety-critical systems like autonomous driving \cite{sun2020scalability,chang2019argoverse}, which rely on LiDAR-based perception, security assessments must be conducted within millisecond-level windows. A reduced query budget directly translates to faster execution, which is vital for evaluating the {instantaneous robustness} of perception models under dynamic, rapidly changing conditions. Our approach enables effective vulnerability detection without exhaustive computational or energy consumption, making it a highly practical proxy for assessing the security of 3D sensors in embedded and real-time applications.}

\noindent \textbf{Visualization Results.}
We provide visualization on adversarial samples generated by our attack, SI-Adv$^w$ \cite{huang2022shape} (white box attack), SI-Adv$^b$ \cite{huang2022shape} (black box attack) and 3DHacker \cite{tao20233dhacker} on the PointNet model in Figure \ref{fig:result}.
We observe that our generated adversarial point clouds exhibit similar geometric structures to their corresponding benign point clouds, \textit{i.e.}, the attacks are quite imperceptible to humans. 
Besides, our adversarial examples have no outliers or uneven point distributions in the local area compared to SI-Adv and 3DHacker. This validates that our hard-label black-box attack can still produce high-quality adversarial samples compared to previous white- and black-box attacks, and is able to alleviate the outlier point problems and produce more imperceptible adversarial samples.

We also provide the results of adversarial samples generated by different fusion methods in Figure~\ref{fig:fusion}. Here, we investigate the effectiveness of our proposed learnable fusion strategy. From this figure, we observe that fixed fusion rate \cite{tao20233dhacker} and random fusion rate result in low-quality adversarial examples with geometric distortion and outliers. Compared to them, learnable fusion leads to better attack performance by preserving more geometric characteristics of the source cloud, demonstrating the effectiveness of the learnable fusion strategy. 
Moreover, since the low-frequency components of an object characterize the basic shape information while the high-frequency components encode fine details and noise, the learnable fusion performance with the low-frequency constraint achieves higher imperceptibility by limiting the noise within high-frequency components.

\subsection{Analysis on Robustness of Our Hard-Label Attack}
To evaluate the robustness of our proposed attack against different adversarial defenses, we conduct experiments on \new{five} widely used defense methods: Lattice Point Classifier (LPC) \cite{li2022robust}, Statistical Outlier Removal (SOR) \cite{zhou2019dup}, Simple Random Sampling (SRS) \cite{yang2019adversarial}\new{, denoising-based defenses UPP \cite{ai2025upp} and N2S3D \cite{li2025learning}}. In particular, following the defense experiments setting on SRS in \cite{huang2022shape}, we conduct SRS by randomly dropping 30\% and 50\% of input points respectively. As shown in Table~\ref{tab:defense}, (1) As for the LPC defense, our proposed attack achieves a better attack performance than SI-Adv$^b$ \cite{huang2022shape} and 3DHacker \cite{tao20233dhacker} in almost all metrics as we generate the adversarial sample with high similarity to the original one in both geometric topology and local point distributions.
(2) We also find that our attack can achieve a higher attack success rate than SI-Adv$^b$ and 3DHacker when attacking the model protected by SOR. This is because our method alleviates the outlier point problems and selects the best adversarial samples with the smallest perturbations, while SI-Adv$^b$ still suffers from the perturbed point of outlier in the sharp component and 3DHacker is limited to the coarsely generated shape due to the fixed fusion.
(3) As for SRS defense, our attack achieves a better attack performance than the compared methods as we generate the adversarial sample with high similarity to the original one in both geometric topology and local point distributions.
\textit{Since SI-Adv utilizes contextual model details to carefully design the perturbations, their $D_c$ metrics are relatively smaller. However, our attack still achieves very competitive $D_c$ while achieving much smaller $D_h$ and $D_{norm}$ compared to SI-Adv.}
\new{(4) As for denoising-based defenses UPP and N2S3D, our attack still shows better performance than other attacks against them, indicating our robustness.}
Overall, our attack is much more robust to existing defense strategies, thus demonstrating its strength.

\begin{table}[t!]
\centering
\caption{\new{Comprehensive evaluation of query efficiency across various victim models. Query costs are decomposed into the Initial Boundary Search (Init. Q) and Iterative Optimization (Opt. Q) stages.}}
\label{tab:comprehensive_efficiency}
\begin{tabular}{l|c|cc|c}
\hline
\multirow{2}{*}{\new{Victim Model}} & \multirow{2}{*}{\new{ASR (\%)}} & \multicolumn{3}{c}{\new{Query Efficiency (Avg. Count)}} \\ \cline{3-5} 
                                       &                                    & \new{Init. Q}   & \new{Opt. Q}   & \new{Total Q}   \\ \hline
\new{PointNet}                    & \new{100}                          & \new{124}       & \new{2281}    & \new{2405}     \\
\new{PointNet++}                  & \new{100}                          & \new{156}       & \new{2644}    & \new{2800}     \\
\new{DGCNN}                       & \new{100}                          & \new{142}       & \new{2518}    & \new{2660}     \\
\hline
\end{tabular}
\end{table}

\begin{table}[t!]
\centering
\caption{\new{Analysis of the relationship between query budget and attack quality on PointNet.}}
\label{tab:budget_quality_tradeoff}
\begin{tabular}{c|c|cc}
\hline
\new{Query Budget} & \new{ASR (\%)} & \new{Hausdorff ($D_h$)} & \new{Chamfer ($D_c$)} \\ \hline
\new{500}          & \new{82.4}  & \new{0.0354}                  & \new{0.0058}                \\
\new{1000}        & \new{94.2}  & \new{0.0216}                  & \new{0.0024}                \\
\new{2000}        & \new{100} & \new{0.0148}                  & \new{0.0016}                \\
\new{2405}        & \new{100} & \new{0.0123}                  & \new{0.0011}                \\
\new{4000}        & \new{100} & \new{0.0105}                  & \new{0.0009}                \\ \hline
\end{tabular}
\end{table}

\begin{table}[t!]
\caption{\new{Comparative results using more local metrics on PointNet.}}
\centering
\setlength{\tabcolsep}{1.4mm}{
\begin{tabular}{c|c|c|ccc}
\hline
\multirow{2}*{\new{Setting}} & \multirow{2}*{\new{Attack}} & \multirow{2}*{\new{ASR(\%)}} & \new{GR} & \new{Curv} & \new{EMD} \\
~ & ~ & ~ & ~ & \new{($10^{-2}$)} & \new{($10^{-2}$)}\\
\hline
\multirow{7}*{\new{White-Box}}
&\new{FGSM \cite{yang2019adversarial}}&\new{100} & \new{0.524} & \new{6.120} & \new{6.842} \\
&\new{PGD \cite{madry2017towards}}&\new{100} & \new{0.406} & \new{5.681} & \new{5.705} \\
&\new{AdvPC \cite{hamdi2020advpc}}&\new{100} & \new{0.325} & \new{1.450} & \new{3.124} \\
&\new{3D-ADV$^p$ \cite{xiang2019generating}}&\new{100} & \new{0.388} & \new{4.210} & \new{4.850} \\
&\new{LG-GAN \cite{zhou2020lg}}&\new{100} & \new{0.284} & \new{0.842} & \new{2.650} \\
&\new{GeoA$^3$ \cite{wen2020geometry}}&\new{100} & \new{0.215} & \new{0.348} & \new{2.415} \\
&\new{SI-Adv$^w$ \cite{huang2022shape}}&\new{100} & \new{0.183} & \new{0.276} & \new{0.783} \\
\hline
\new{Black-Box} & \new{SI-Adv$^b$ \cite{huang2022shape}} & \new{100} & \new{0.205} & \new{0.312} & \new{0.942} \\
\hline
\multirow{2}*{\tabincell{c}{\new{Hard-Label}\\ \new{Black-Box}}}
&\new{3DHacker \cite{tao20233dhacker}} & \new{100} & \new{0.224} & \new{0.385} & \new{0.950} \\
~ & \new{Ours} & \new{100} & \new{\textbf{0.192}} & \new{\textbf{0.284}} & \new{\textbf{0.811}} \\
\hline
\end{tabular}}
\label{tab:more_metric}
\end{table}

\noindent \textbf{\new{Evaluations on Local Preservation.}}
\new{We further report more local structure metrics: surface curvature (Curv) \cite{wen2020geometry}, geometric regularity (GR) \cite{wen2020geometry}, and earth mover’s distance (EMD) for evaluations in Table~\ref{tab:more_metric}. It shows that our attack also achieves great local structure preservation.}

\begin{table*}[t!]
\caption{Main ablation study on different components of the proposed hard-label black-box attack method.}
\centering
\setlength{\tabcolsep}{1.5mm}{
\begin{tabular}{c|ccccc|cccc|cccc}
\hline~ &
\multicolumn{2}{c}{Boundary Cloud Generation} & &\multicolumn{2}{c}{Boundary Cloud Optimization} &\multicolumn{4}{|c}{PointNet \cite{qi2017pointnet}} & \multicolumn{4}{|c}{PAConv \cite{xu2021paconv}}\\
\cline{2-3} \cline{5-14}
~ & Spectrum & Learnable & & Joint & Geometric & \multirow{2}*{ASR(\%)} &\multirow{2}*{$D_h$} & \multirow{2}*{$D_c$} & \multirow{2}*{$D_{norm}$} & \multirow{2}*{ASR(\%)} &\multirow{2}*{$D_h$} & \multirow{2}*{$D_c$} & \multirow{2}*{$D_{norm}$}   \\
~ & Guidance & Fusion & & Optimization & Walking & & & & & & & & 
\\
\hline
1. & $\times$ & $\times$ & & $\times$ & $\times$ & 82.8 & 0.0634 & 0.0143 & 2.3117 & 79.4 & 0.0413 & 0.0107 & 1.8584\\
2. & $\checkmark$ & $\times$ & & $\times$ & $\times$ & 83.5 & 0.0274 & 0.0094 & 1.2496 & 72.9 & 0.0098 & 0.0055 & 1.3108\\
3. & $\checkmark$ & $\times$ & & $\checkmark$ & $\times$ & 92.7 & 0.0144 & 0.0024 & 1.0517 & 94.1 & 0.0056 & 0.0029 & 1.1033\\
4. & $\checkmark$ & $\checkmark$ & & $\times$ & $\times$ & 95.1 & 0.0207 & 0.0065 & 1.0741 & 97.4 & 0.0073 & 0.0043 & 1.0217\\
5. & $\checkmark$ & $\checkmark$ & & $\checkmark$ & $\times$ & \textbf{100}  & 0.0141 & 0.0018 & 0.9243 & \textbf{100}  & 0.0042 & 0.0023 & 0.9931\\
6. & $\checkmark$ & $\times$ & & $\checkmark$ & $\checkmark$ & \textbf{100} & 0.0155 & 0.0023 & 0.8973 & \textbf{100} & 0.0051 & 0.0030 & 0.9393\\
7. & $\checkmark$ & $\checkmark$ & & $\checkmark$ & $\checkmark$ & \textbf{100} & \textbf{0.0123} & \textbf{0.0011} & \textbf{0.8112} & \textbf{100} & \textbf{0.0037} & \textbf{0.0012} & \textbf{0.8579}\\
\hline
\end{tabular}}
\label{tab:ablation1}
\end{table*}

\subsection{\new{Analysis of Query Cost}}
\new{We also provide a comprehensive and systematic analysis of query efficiency from multiple perspectives, including average query cost, stage-wise query decomposition, and the relationship between query budget and attack quality.
As shown in Table~\ref{tab:comprehensive_efficiency}, our method achieves a low average query cost across different victim models. These results demonstrate that our method can consistently achieve highly effective attacks with a relatively small number of queries, highlighting its practicality in hard-label black-box settings.
We further decompose the total query cost into two stages: the Initial Boundary Search (Init. Q) and Iterative Optimization (Opt. Q). The results show that the initialization stage requires only a small fraction of the total queries, while the majority of queries are spent on iterative optimization.
To further analyze the efficiency of query utilization, we report the relationship between query budget and attack performance in Table~\ref{tab:budget_quality_tradeoff}. 
As the query budget increases, both Hausdorff distance and Chamfer distance consistently decrease, showing that additional queries are effectively used to refine perturbations rather than merely increasing the success rate. Average query 2405 is effective enough to achieve our best performance by balancing the performance and efficiency.}

\subsection{Ablation Study}
\subsubsection{Main ablation}
To analyze how each component contributes to the whole attack method, we conduct a main ablation study to validate the effectiveness of different components (\textit{i.e.}, boundary cloud generation and optimization) on both PointNet and PAConv victim models, as shown in Table~\ref{tab:ablation1}.
We start from the baseline model (line 1) which does not utilize learnable spectrum fusion and the geometric spectrum walking strategies. 
Instead, the baseline model solely utilizes coordinate fusion and walking strategies with fixed fusion rates to directly generate the adversarial point clouds. Experimental results show that the baseline model achieves low attack success rates and large perturbation sizes.
By applying the spectrum fusion (line 2) to the baseline model, the attack performance increases a lot, demonstrating that spectrum fusion is able to alleviate the geometric distortion compared to coordinate fusion.
By further applying the learnable fusion strategy and spectrum walking strategies (lines 3 and 4) to the second variant (line 2), the quality of our generated adversarial samples is improved. It validates that the learnable fusion is able to keep better geometric shape of the original point cloud compared to the fixed fusion while the additional spectrum fusion is able to alleviate the local optimum caused by the coordinate fusion.
Lines 5-7 also demonstrate the effectiveness of our newly designed geometric-aware walking strategy for optimizing the boundary cloud along the decision boundary.
Overall, the whole framework with learnable spectrum fusion and the geometry-aware joint coordinate-spectrum optimization strategies achieves the best performance.

\subsubsection{Ablation on the boundary cloud generation}
\noindent \textbf{Investigation on different fusion strategies.}
To verify the effects of our proposed spectrum fusion method in the boundary cloud generation stage, we conduct the experiments by replacing the spectrum fusion method with different strategies while maintaining the latter procedure and settings in the boundary cloud optimization stage the same. Specifically, two general strategies are compared: traditional coordinate fusion (which fuses the source-cloud and target-cloud in the coordinate space with proper fusion rate) and simple random perturbation (which directly adds point-wise noise to the source-cloud to reach the decision boundary). As shown in Table~\ref{tab:ablation2}, our spectrum fusion achieves the smallest perturbations than other strategies in all metrics. 
This is because: (1) coordinate fusion will destroy the geometric structure by averaging different shapes of 3D objects; (2) random perturbation will lead to outliers and uneven point distribution without geometric awareness. 

\begin{table}[t!]
\caption{Ablation on the boundary cloud generation on the PointNet model.}
\centering
\setlength{\tabcolsep}{1.0mm}{
\begin{tabular}{c|c|cccc}
\hline
Component & Variants & ASR(\%) &$D_h$ & $D_c$ & $D_{norm}$\\
\hline
\multirow{3}*{\tabincell{c}{Fusion\\ Strategy}} & Spectrum Fusion & \textbf{100} & \textbf{0.0123} & \textbf{0.0011} & \textbf{0.8112} \\
~ & Coordinate Fusion &82.4 &0.0354 &0.0103 & 1.8373\\ 
~ & Random Perturbation &86.1 & 0.0371 &0.0079 & 1.6417\\
\hline
\multirow{3}*{\tabincell{c}{Fusion\\ Weight}} & Learnable Weight+Bank & \textbf{100} & 0.0123 & 0.0011 & \textbf{0.8112}\\
~ & Learnable Weight &\textbf{100} & \textbf{0.0117}&\textbf{0.0010}&0.8154\\
~ & Fixed Weight & 92.7 & 0.0137 & 0.0022 &1.0157\\
\hline
\multirow{3}*{\tabincell{c}{Training\\ of Fusion \\ Module}} &  Instance-guided & \textbf{100} & \textbf{0.0123} & \textbf{0.0011} & \textbf{0.8112}  \\
~ & Class-guided &\textbf{100} & 0.0128&0.0014&0.8231\\
~ & Dataset-guided &96.7& 0.0134 &0.0023 &0.9841\\
\hline
\end{tabular}}
\label{tab:ablation2}
\end{table}

\begin{table}[t!]
\caption{Analysis of the fusion weight bank on the PointNet model. ``Time" denotes the averaged time for generating one sample, ``Memory" means the GPU memory cost.}
\centering
\setlength{\tabcolsep}{1.4mm}{
\begin{tabular}{c|ccc|c|c}
\hline
Variants &$D_h$ & $D_c$ & $D_{norm}$ & Time$\downarrow$ & Memory$\downarrow$\\
\hline
with bank & 0.0123 & 0.0011 & 0.8112 &\textbf{2.42}s&\textbf{14791}M \\
without bank & 0.0117 & 0.0010 & 0.8154 &14.67s&23018M
\\
\hline
\end{tabular}}
\label{tab:analysis1}
\end{table}

\begin{table}[t!]
\caption{\new{Ablation study on different strategies for fusion weight $\alpha$ selection on PointNet. ``A.Q.'' denotes the average queries required to reach 100\% ASR or the maximum iteration limit.}}
\centering
\setlength{\tabcolsep}{1.2mm}{
\begin{tabular}{c|cccc|c}
\hline
\new{Strategy} & \new{ASR(\%)} & \new{$D_h$} & \new{$D_c$} & \new{$D_{norm}$} & \new{A.Q. $\downarrow$} \\
\hline
\new{Random Sampling}  & \new{86.4} & \new{0.0382} & \new{0.0081} & \new{1.6842} & \new{10540} \\
\new{Grid Search}  & \new{93.1} & \new{0.0132} & \new{0.0020} & \new{1.0148} & \new{7812} \\
\hline
\new{\textbf{Ours (Discriminator)}} & \new{\textbf{100}} & \new{\textbf{0.0123}} & \new{\textbf{0.0011}} & \new{\textbf{0.8112}} & \new{\textbf{2405}} \\
\hline
\end{tabular}}
\label{tab:ablation_alpha}
\end{table}

\noindent \textbf{Investigation on different designs of fusion weights.}
We also investigate the effectiveness of the design of our learnable fusion weights. As shown in Table~\ref{tab:ablation2}, directly utilizing fixed fusion weight results in worse attack performance with larger perturbation size. This is because the fixed weight is not suitable for all possible fusion cases of different object classes. By replacing the fixed fusion weight with the learnable one, our attack performance increases a lot. We observe that the ``Learnable Weight+Bank" variant achieves very similar performance as the ``Learnable Weight" variant, where the former collects the learnable weights of each specific class-guided fusion case during the training and utilizes them from the bank during the inference while the latter predicts the learnable weights in both training and inference stages. This demonstrates that the fusion weights of the same class share similar characteristics, which represent the same class-sensitive tolerance to the noise when we add the target clouds to this class-specific point cloud. Although directly utilizing ``Learnable Weight" can achieve more fine-grained performance, it will cost much time and memory during the inference, as shown in Table~\ref{tab:analysis1}. Therefore, we utilize the ``Learnable Weight+Bank" variant as our pipeline.

\begin{figure*}[!t]
	\centering
	\includegraphics[width=0.9\textwidth]{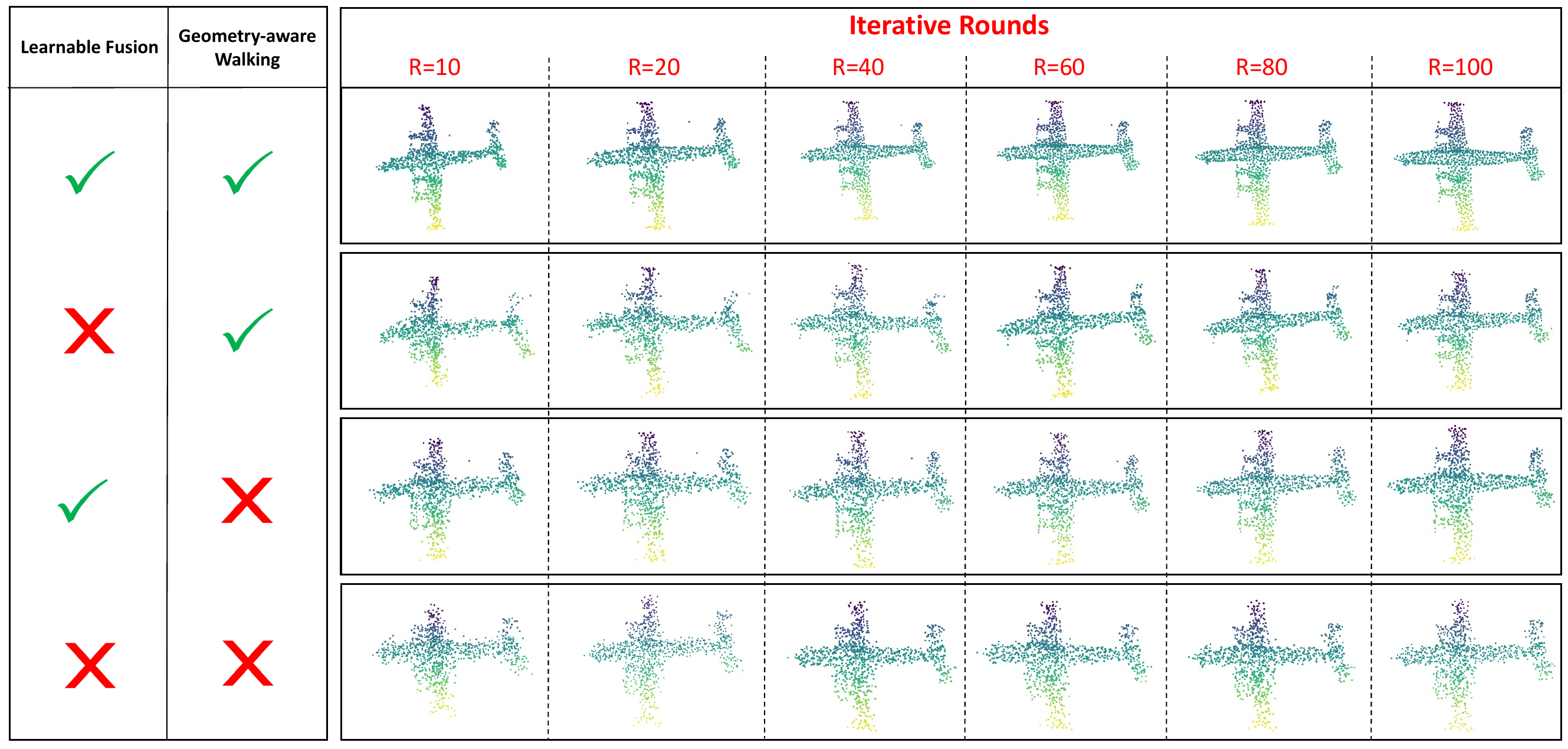}
	\caption{Visualization results of adversarial samples generated by different optimization steps.}
	\label{fig:iteration}
\end{figure*}

\noindent \textbf{Investigation on different training principles of fusion module.}
At last, we investigate the effect of different training principles of the fusion module. To be specific, we conduct three ablation variants to generate different kinds of learnable fusion weights, as shown in Table~\ref{tab:ablation2}.
The ``instance-guided" variant means that we generate the fusion weights of each point-cloud pair during the training and randomly sample one of them to fuse the point-cloud pair of the same class during the inference.
The ``class-guided" variant means that we generate the fusion weight for the entire class samples.
The ``datasets-guided" variant means that we generate a fusion weight for all point-cloud pairs without considering their classes.
Results show that the ``datasets-guided" variant achieves the worst performance as it is similar to the strategy of fixed fusion weight.
The ``instance-guided" variant achieves better attack performance than the ``class-guided" variant since it collects more specific fusion cases of the same class objects.
\new{Table~\ref{tab:ablation_alpha} also demonstrates that our training process of the discriminator is essential, as it is more explicit and effective than grid search and random sampling to learn adaptive and appropriate weights for efficiently generating high-imperceptible boundary point clouds.}

\begin{table}[t!]
\caption{Ablation on the boundary cloud optimization on the PointNet model.}
\centering
\setlength{\tabcolsep}{1.0mm}{
\begin{tabular}{c|c|cccc}
\hline
Component & Variants & ASR(\%) &$D_h$ & $D_c$ & $D_{norm}$\\
\hline
\multirow{3}*{\tabincell{c}{Walking\\ Strategy}} & Joint Walking &  \textbf{100} & \textbf{0.0123} & \textbf{0.0011} & \textbf{0.8112}   \\
~ & only Coordinate &93.1&0.0253&0.0023&1.0784\\
~ & only Spectrum &90.7&0.0384&0.0038&1.7902\\
\hline
\multirow{2}*{\tabincell{c}{Optimization\\ Strategy}} & Geometry-aware & \textbf{100} & \textbf{0.0123} & \textbf{0.0011} & \textbf{0.8112}  \\
~ & Binary Search & \textbf{100} & 0.0141 & 0.0018 & 0.9243 \\
\hline
\multirow{4}*{\tabincell{c}{Iteration\\ Round}} & \textit{R}=30 &\textbf{100}&0.0273&0.0035&1.2807\\
~ & \textit{R}=70 &\textbf{100}&0.0139&0.0015&0.8710\\
~ & \textit{R}=100 &  \textbf{100} & 0.0123 & \textbf{0.0011} & 0.8112\\
& \textit{R}=130& \textbf{100} & \textbf{0.0121} & \textbf{0.0011} & \textbf{0.8079} \\
\hline
\end{tabular}}
\label{tab:ablation3}
\end{table}

\begin{table}[t!]
\caption{Analysis of the geometry-aware optimization on the PointNet model.}
\centering
\setlength{\tabcolsep}{1.0mm}{
\begin{tabular}{c|c|c|c}
\hline
Variants &Query Number$\downarrow$ & Iteration Step$\downarrow$ & Time$\downarrow$ \\
\hline
Geometry-aware &\textbf{2405}&\textbf{100}&\textbf{2.42s}\\
Binary Search &13392&200&21.8s
\\
\hline
\end{tabular}}
\label{tab:analysis2}
\end{table}

\subsubsection{Ablation on the boundary cloud optimization}
\noindent \textbf{Investigation on different walking strategies.}
In the boundary cloud optimization stage, we design a spectrum walking method in addition to the coordinate one to jump out of the local optimum. To investigate the effect of this spectrum walking strategy, as shown in Table~\ref{tab:ablation3}, we remove one of them to conduct the ablations for comparison. From this table, without spectrum walking, the attack process is easily trapped into the local optimum, resulting in larger perturbations. Without coordinate walking, it is hard to constrain the point-wise imperceptibility during the optimization process, thus achieving the worst performance. By utilizing both of them, our model can preserve both high imperceptibility and geometric smoothness. This further demonstrates the effectiveness of our proposed joint coordinate-spectrum walking strategy.

\noindent \textbf{Investigation on different optimization strategies.}
We also investigate the effectiveness of our proposed geometry-aware optimization strategy. We compared our geometry-aware strategy with the traditional normal-vector-based binary search strategy in Table~\ref{tab:ablation3}. We find that our optimization strategy is able to achieve better attack performance (\textit{i.e.}, lower perturbation size). 
The reason is: As for the binary search strategy, due to the limited query budget and the non-linearity of the boundary, the estimated normal vector may be inaccurate and result in wrong predictions. Instead, our geometry-aware optimization strategy is to adjust the boundary cloud along a semicircular path based on the accurate curvature contexts of the surface. Therefore, the boundary cloud can be better optimized to achieve high quality. Moreover, as shown in Table~\ref{tab:analysis2}, estimating normal vectors is
time-consuming as it needs to generate multiple queries.
In comparison, our strategy solely relies on the geometric curvature contexts for optimizing the point cloud along the decision boundary, significantly increasing the query efficiency.

\noindent \textbf{Sensitivity on the iteration rounds.}
As shown in Table~\ref{tab:ablation3}, we further conduct the ablation on the iteration rounds of the boundary cloud optimization. Our model achieves the best performance when the iteration step $R$ is set to 130. However, the model with $R=130$ is slightly better than the model with $R=100$, but results in much more time consumption. To balance both the performance and time cost, we choose $R = 100$ in all our experiments. Visualizations on adversarial point clouds generated by variants of different optimization steps are shown in Figure~\ref{fig:iteration}.

\begin{table}[t!]
\caption{\new{Sensitivity analysis on the key parameters of KNN graph construction and spectral energy split on the PointNet model.}}
\centering
\setlength{\tabcolsep}{1.2mm}{
\begin{tabular}{c|c|cccc}
\hline
\new{Parameter} & \new{Variants} & \new{ASR(\%)} & \new{$D_h$} & \new{$D_c$} & \new{$D_{norm}$} \\
\hline
\multirow{4}*{\new{\tabincell{c}{Number of \\ Neighbors $K$}}} 
& \new{$K=5$} & \new{98.2} & \new{0.0145} & \new{0.0018} & \new{0.9324} \\
~ & \new{\textbf{$K=10$}} & \new{\textbf{100}} & \new{\textbf{0.0123}} & \new{\textbf{0.0011}} & \new{\textbf{0.8112}} \\
~ & \new{$K=20$} & \new{\textbf{100}} & \new{0.0128} & \new{0.0013} & \new{0.8245} \\
~ & \new{$K=30$} & \new{\textbf{100}} & \new{0.0135} & \new{0.0016} & \new{0.8510} \\
\hline
\multirow{5}*{\new{\tabincell{c}{Spectral \\ Energy \\ Split}}} 
& \new{\textbf{(75\%, 90\%)}} & \new{\textbf{100}} & \new{\textbf{0.0123}} & \new{\textbf{0.0011}} & \new{\textbf{0.8112}} \\
~ & \new{(73\%, 90\%)} & \new{\textbf{100}} & \new{0.0128} & \new{0.0011} & \new{0.8153} \\
~ & \new{(77\%, 90\%)} & \new{\textbf{100}} & \new{0.0121} & \new{0.0011} & \new{0.8136} \\
~ & \new{(75\%, 88\%)} & \new{\textbf{100}} & \new{0.0126} & \new{0.0009} & \new{0.8118} \\
~ & \new{(75\%, 92\%)} & \new{\textbf{100}} & \new{0.0142} & \new{0.0016} & \new{0.8164} \\
\hline
\end{tabular}}
\label{tab:sensitivity_analysis}
\end{table}

\subsubsection{\new{Sensitive to other hyperparameters}} 
\new{
To further investigate the design choice of the KNN graph construction and the frequency band, we conduct a sensitivity analysis on the number of neighbors $K$ in KNN graph construction and the spectral energy split ratios in spectral decomposition, as shown in Table~\ref{tab:sensitivity_analysis}.
(1) \textit{Number of Neighbors $K$}: We evaluate $K$ values ranging from 5 to 30. Our method achieves the best overall performance at $K=10$, which provides a stable local graph structure for geometry-aware gradient estimation, demonstrating the robustness of our graph-based representation.
(2) \textit{Spectral Energy Split}: Previous works \cite{liu2022point} find that point clouds have almost 75\% of energy within the lowest 100 frequencies and almost 90\% of energy within the lowest 400 frequencies. 
By testing various low-and-high frequency split ratios, following \cite{liu2022point}, we confirm that the setting of (75\%, 90\%) serves as an optimal configuration for spectral decomposition. It consistently yields the most balanced results across $D_h$ and $D_c$, effectively capturing the essential manifold features for black-box optimization. Slight variations in these ratios do not lead to significant performance degradation, which validates that our spectral guidance is robust and generalizes well across different frequency boundary definitions.}

\subsection{Other ablations}
\begin{figure}[!t]
	\centering
	\includegraphics[width=0.49\textwidth]{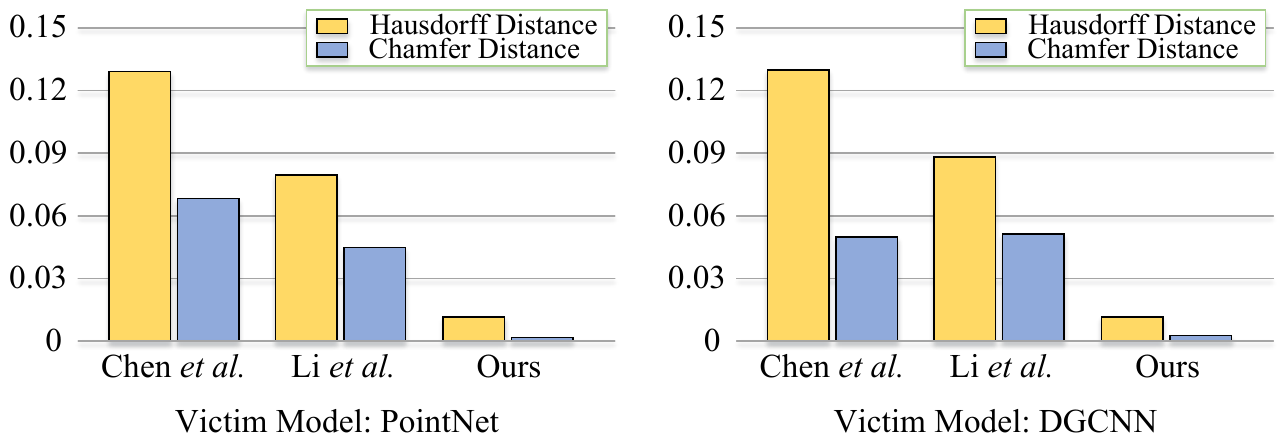}
	\caption{Comparison on the 2D hard-label setting.}
	\label{fig:decision}
\end{figure}

\noindent \textbf{Effect of our 3D hard-label pipeline.}
Since existing 3D attacks rely on either model parameters or output logits, they cannot be adapted to hard-label setting for comparison. Therefore, we re-implement two 2D hard-label settings (\textit{i.e.}, Chen \textit{et al.} \cite{chen2020hopskipjumpattack} and Li \textit{et al.} \cite{li2021nonlinear}) into 3D domain to evaluate our 3D hard-label pipeline. As shown in Figure~\ref{fig:decision}, our method performs much better, demonstrating that directly applying the 2D decision boundary mechanism to the 3D domain is not effective.

\begin{figure}[!t]
	\centering
	\includegraphics[width=0.49\textwidth]{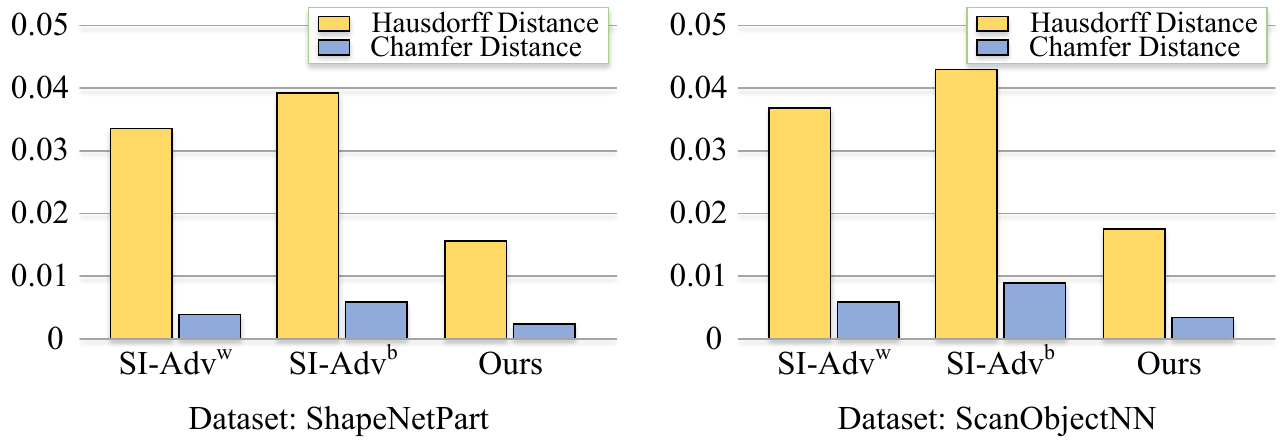}
	\caption{3D attack comparison on more 3D datasets.}
	\label{fig:dataset}
\end{figure}

\noindent \textbf{Comparison on more datasets.}
To further validate the robustness of our proposed attack method, we implement the compared methods on more 3D datasets (\textit{i.e.}, ShapeNetPart and ScanObjectNN \cite{uy2019revisiting}). As shown in \new{Figure}~\ref{fig:dataset}, our method achieves better attack performance than the SOTA methods, demonstrating its effectiveness.

\section{Conclusion}
In this paper, we introduce a new yet challenging 3D attack setting, \textit{i.e.}, attacking point clouds with black-box hard labels. To address this practical setting, we propose a novel attack method based on our newly designed decision boundary algorithm, which adopts learnable spectrum fusion to generate boundary clouds with high imperceptibility and employs a joint coordinate-spectrum walking strategy to move the boundary clouds along the decision boundary for further optimization with trivial perturbations. During the optimization process, we also propose to update the boundary cloud by solely considering the geometric curvature information of the decision boundary. 
Experimental results demonstrate the vulnerability of popular 3D models to our proposed attack and validate the robustness of our adversarial point clouds.

\bibliographystyle{IEEEtran}
\bibliography{reference,reference2}

\begin{thebibliography}{10}
\providecommand{\url}[1]{#1}
\csname url@samestyle\endcsname
\providecommand{\newblock}{\relax}
\providecommand{\bibinfo}[2]{#2}
\providecommand{\BIBentrySTDinterwordspacing}{\spaceskip=0pt\relax}
\providecommand{\BIBentryALTinterwordstretchfactor}{4}
\providecommand{\BIBentryALTinterwordspacing}{\spaceskip=\fontdimen2\font plus
\BIBentryALTinterwordstretchfactor\fontdimen3\font minus \fontdimen4\font\relax}
\providecommand{\BIBforeignlanguage}[2]{{%
\expandafter\ifx\csname l@#1\endcsname\relax
\typeout{** WARNING: IEEEtran.bst: No hyphenation pattern has been}%
\typeout{** loaded for the language `#1'. Using the pattern for}%
\typeout{** the default language instead.}%
\else
\language=\csname l@#1\endcsname
\fi
#2}}
\providecommand{\BIBdecl}{\relax}
\BIBdecl

\bibitem{madry2017towards}
A.~Madry, A.~Makelov, L.~Schmidt, D.~Tsipras, and A.~Vladu, ``Towards deep learning models resistant to adversarial attacks,'' in \emph{Proceedings of the International Conference on Learning Representations (ICLR)}, 2017.

\bibitem{tu2019autozoom}
C.-C. Tu, P.~Ting, P.-Y. Chen, S.~Liu, H.~Zhang, J.~Yi, C.-J. Hsieh, and S.-M. Cheng, ``Autozoom: Autoencoder-based zeroth order optimization method for attacking black-box neural networks,'' in \emph{Proceedings of the AAAI Conference on Artificial Intelligence}, vol.~33, no.~01, 2019, pp. 742--749.

\bibitem{zhang2019eye}
H.~Zhang, G.~Wang, Z.~Lei, and J.-N. Hwang, ``Eye in the sky: Drone-based object tracking and 3d localization,'' in \emph{Proceedings of the 27th ACM international conference on multimedia}, 2019, pp. 899--907.

\bibitem{zhao2025lossless}
R.~Zhao, Y.~Zhang, J.~Mou, W.~Puech, and J.~Weng, ``Lossless and universal 3d object encryption with differentiated visual effects upon decryption: A novel paradigm,'' \emph{IEEE Transactions on Dependable and Secure Computing}, 2025.

\bibitem{zhong2020reliable}
B.~Zhong, H.~Huang, and E.~Lobaton, ``Reliable vision-based grasping target recognition for upper limb prostheses,'' \emph{IEEE Transactions on Cybernetics}, 2020.

\bibitem{singh20203d}
S.~P. Singh, L.~Wang, S.~Gupta, H.~Goli, P.~Padmanabhan, and B.~Guly{\'a}s, ``3d deep learning on medical images: a review,'' \emph{Sensors}, vol.~20, no.~18, p. 5097, 2020.

\bibitem{tsai2020robust}
T.~Tsai, K.~Yang, T.-Y. Ho, and Y.~Jin, ``Robust adversarial objects against deep learning models,'' in \emph{Proceedings of the AAAI Conference on Artificial Intelligence}, vol.~34, no.~01, 2020, pp. 954--962.

\bibitem{zhao2020isometry}
Y.~Zhao, Y.~Wu, C.~Chen, and A.~Lim, ``On isometry robustness of deep 3d point cloud models under adversarial attacks,'' in \emph{Proceedings of the IEEE Conference on Computer Vision and Pattern Recognition (CVPR)}, 2020, pp. 1201--1210.

\bibitem{zhou2020lg}
H.~Zhou, D.~Chen, J.~Liao, K.~Chen, X.~Dong, K.~Liu, W.~Zhang, G.~Hua, and N.~Yu, ``Lg-gan: Label guided adversarial network for flexible targeted attack of point cloud based deep networks,'' in \emph{Proceedings of the IEEE Conference on Computer Vision and Pattern Recognition (CVPR)}, 2020, pp. 10\,356--10\,365.

\bibitem{wen2020geometry}
Y.~Wen, J.~Lin, K.~Chen, C.~P. Chen, and K.~Jia, ``Geometry-aware generation of adversarial point clouds,'' \emph{IEEE Transactions on Pattern Analysis and Machine Intelligence (TPAMI)}, 2020.

\bibitem{xiao2025dual}
X.~Xiao, Y.~Zhang, R.~Zhao, Z.~Hua, W.~Wen, and Y.~Fang, ``A dual-protection method for 3d object security and copyright: Watermark embedding during decryption,'' \emph{IEEE Transactions on Dependable and Secure Computing}, 2025.

\bibitem{xie2022stealthy}
S.~Xie, Y.~Yan, and Y.~Hong, ``Stealthy 3d poisoning attack on video recognition models,'' \emph{IEEE Transactions on Dependable and Secure Computing}, vol.~20, no.~2, pp. 1730--1743, 2022.

\bibitem{xiang2019generating}
C.~Xiang, C.~R. Qi, and B.~Li, ``Generating 3d adversarial point clouds,'' in \emph{Proceedings of the IEEE Conference on Computer Vision and Pattern Recognition (CVPR)}, 2019, pp. 9136--9144.

\bibitem{huang2022shape}
Q.~Huang, X.~Dong, D.~Chen, H.~Zhou, W.~Zhang, and N.~Yu, ``Shape-invariant 3d adversarial point clouds,'' in \emph{Proceedings of the IEEE/CVF Conference on Computer Vision and Pattern Recognition}, 2022, pp. 15\,335--15\,344.

\bibitem{li2020qeba}
H.~Li, X.~Xu, X.~Zhang, S.~Yang, and B.~Li, ``Qeba: Query-efficient boundary-based blackbox attack,'' in \emph{Proceedings of the IEEE/CVF conference on computer vision and pattern recognition}, 2020, pp. 1221--1230.

\bibitem{li2022decision}
X.-C. Li, X.-Y. Zhang, F.~Yin, and C.-L. Liu, ``Decision-based adversarial attack with frequency mixup,'' \emph{IEEE Transactions on Information Forensics and Security}, vol.~17, pp. 1038--1052, 2022.

\bibitem{li2021nonlinear}
H.~Li, L.~Li, X.~Xu, X.~Zhang, S.~Yang, and B.~Li, ``Nonlinear projection based gradient estimation for query efficient blackbox attacks,'' in \emph{International Conference on Artificial Intelligence and Statistics}.\hskip 1em plus 0.5em minus 0.4em\relax PMLR, 2021, pp. 3142--3150.

\bibitem{lu2021large}
S.-P. Lu, R.~Wang, T.~Zhong, and P.~L. Rosin, ``Large-capacity image steganography based on invertible neural networks,'' in \emph{Proceedings of the IEEE/CVF conference on computer vision and pattern recognition}, 2021, pp. 10\,816--10\,825.

\bibitem{xu2022robust}
Y.~Xu, C.~Mou, Y.~Hu, J.~Xie, and J.~Zhang, ``Robust invertible image steganography,'' in \emph{Proceedings of the IEEE/CVF Conference on Computer Vision and Pattern Recognition}, 2022, pp. 7875--7884.

\bibitem{hu2021graph}
W.~Hu, J.~Pang, X.~Liu, D.~Tian, C.-W. Lin, and A.~Vetro, ``Graph signal processing for geometric data and beyond: Theory and applications,'' \emph{IEEE Transactions on Multimedia}, 2021.

\bibitem{hu2022exploring}
Q.~Hu, D.~Liu, and W.~Hu, ``Exploring the devil in graph spectral domain for 3d point cloud attacks,'' \emph{arXiv preprint arXiv:2202.07261}, 2022.

\bibitem{ortega2022introduction}
A.~Ortega, \emph{Introduction to graph signal processing}.\hskip 1em plus 0.5em minus 0.4em\relax Cambridge University Press, 2022.

\bibitem{tao20233dhacker}
Y.~Tao, D.~Liu, P.~Zhou, Y.~Xie, W.~Du, and W.~Hu, ``3dhacker: Spectrum-based decision boundary generation for hard-label 3d point cloud attack,'' in \emph{Proceedings of the IEEE International Conference on Computer Vision (ICCV)}, 2023.

\bibitem{yu2018multi}
T.~Yu, J.~Meng, and J.~Yuan, ``Multi-view harmonized bilinear network for 3d object recognition,'' in \emph{Proceedings of the IEEE Conference on Computer Vision and Pattern Recognition (CVPR)}, 2018, pp. 186--194.

\bibitem{qi2017pointnet}
C.~R. Qi, H.~Su, K.~Mo, and L.~J. Guibas, ``Pointnet: Deep learning on point sets for 3d classification and segmentation,'' in \emph{Proceedings of the IEEE Conference on Computer Vision and Pattern Recognition (CVPR)}, 2017, pp. 652--660.

\bibitem{qi2017pointnet++}
C.~R. Qi, L.~Yi, H.~Su, and L.~J. Guibas, ``Pointnet++: Deep hierarchical feature learning on point sets in a metric space,'' \emph{Advances in Neural Information Processing Systems (NIPS)}, 2017.

\bibitem{duan2019structural}
Y.~Duan, Y.~Zheng, J.~Lu, J.~Zhou, and Q.~Tian, ``Structural relational reasoning of point clouds,'' in \emph{Proceedings of the IEEE Conference on Computer Vision and Pattern Recognition (CVPR)}, 2019, pp. 949--958.

\bibitem{liu2019densepoint}
Y.~Liu, B.~Fan, G.~Meng, J.~Lu, S.~Xiang, and C.~Pan, ``Densepoint: Learning densely contextual representation for efficient point cloud processing,'' in \emph{Proceedings of the IEEE International Conference on Computer Vision (ICCV)}, 2019, pp. 5239--5248.

\bibitem{yang2019modeling}
J.~Yang, Q.~Zhang, B.~Ni, L.~Li, J.~Liu, M.~Zhou, and Q.~Tian, ``Modeling point clouds with self-attention and gumbel subset sampling,'' in \emph{Proceedings of the IEEE Conference on Computer Vision and Pattern Recognition (CVPR)}, 2019, pp. 3323--3332.

\bibitem{xu2021paconv}
M.~Xu, R.~Ding, H.~Zhao, and X.~Qi, ``Paconv: Position adaptive convolution with dynamic kernel assembling on point clouds,'' in \emph{Proceedings of the IEEE/CVF Conference on Computer Vision and Pattern Recognition}, 2021, pp. 3173--3182.

\bibitem{wicker2019robustness}
M.~Wicker and M.~Kwiatkowska, ``Robustness of 3d deep learning in an adversarial setting,'' in \emph{Proceedings of the IEEE Conference on Computer Vision and Pattern Recognition (CVPR)}, 2019, pp. 11\,767--11\,775.

\bibitem{zheng2019pointcloud}
T.~Zheng, C.~Chen, J.~Yuan, B.~Li, and K.~Ren, ``Pointcloud saliency maps,'' in \emph{Proceedings of the IEEE International Conference on Computer Vision (ICCV)}, 2019, pp. 1598--1606.

\bibitem{zhao2021point}
H.~Zhao, L.~Jiang, J.~Jia, P.~H. Torr, and V.~Koltun, ``Point transformer,'' in \emph{Proceedings of the IEEE/CVF international conference on computer vision}, 2021, pp. 16\,259--16\,268.

\bibitem{yu2022point}
X.~Yu, L.~Tang, Y.~Rao, T.~Huang, J.~Zhou, and J.~Lu, ``Point-bert: Pre-training 3d point cloud transformers with masked point modeling,'' in \emph{Proceedings of the IEEE/CVF Conference on Computer Vision and Pattern Recognition}, 2022, pp. 19\,313--19\,322.

\bibitem{wang2019dynamic}
Y.~Wang, Y.~Sun, Z.~Liu, S.~E. Sarma, M.~M. Bronstein, and J.~M. Solomon, ``Dynamic graph cnn for learning on point clouds,'' \emph{ACM Transactions on Graphics (tog)}, vol.~38, no.~5, pp. 1--12, 2019.

\bibitem{uy2019revisiting}
M.~A. Uy, Q.-H. Pham, B.-S. Hua, T.~Nguyen, and S.-K. Yeung, ``Revisiting point cloud classification: A new benchmark dataset and classification model on real-world data,'' in \emph{Proceedings of the IEEE/CVF international conference on computer vision}, 2019, pp. 1588--1597.

\bibitem{gao2025distributed}
Y.~Gao, S.~Li, T.~Xu, S.~Lakshminarayana, S.~Bu, C.~Gu, and Q.~Ai, ``Distributed model predictive control strategy for multi-energy virtual power plant based on digital twin,'' \emph{IEEE Transactions on Smart Grid}, 2025.

\bibitem{xu2025prediction}
B.~Xu, G.~Rang, R.~Xie, W.~Li, D.~Gong, Z.~Fan, S.~Yang, and J.~He, ``A prediction approach based on long short-term memory networks for dynamic multiobjective optimization,'' \emph{Expert Systems with Applications}, vol. 283, p. 127792, 2025.

\bibitem{fu2026hyperr3snet}
J.~Fu, C.~Wang, M.~Liu, X.~Li, Y.~Liu, W.~Shi, and R.~Wang, ``Hyperr3snet: Leveraging hyperbolic space and vision foundation models for remote sensing semantic segmentation,'' \emph{IEEE Transactions on Geoscience and Remote Sensing}, 2026.

\bibitem{chen2025learning}
P.~Chen, X.~Nie, Y.~Ning, and Y.~Zhang, ``Learning efficient and adaptive cross-channel dependencies for weakly-supervised object detection,'' \emph{IEEE Transactions on Multimedia}, 2025.

\bibitem{zhang2019defense}
Y.~Zhang, G.~Liang, T.~Salem, and N.~Jacobs, ``Defense-pointnet: Protecting pointnet against adversarial attacks,'' in \emph{2019 IEEE International Conference on Big Data (Big Data)}, 2019, pp. 5654--5660.

\bibitem{liu2021imperceptible}
D.~Liu and W.~Hu, ``Imperceptible transfer attack and defense on 3d point cloud classification,'' \emph{arXiv preprint arXiv:2111.10990}, 2021.

\bibitem{kim2021minimal}
J.~Kim, B.-S. Hua, T.~Nguyen, and S.-K. Yeung, ``Minimal adversarial examples for deep learning on 3d point clouds,'' in \emph{Proceedings of the IEEE/CVF International Conference on Computer Vision}, 2021, pp. 7797--7806.

\bibitem{shi2022shape}
Z.~Shi, Z.~Chen, Z.~Xu, W.~Yang, Z.~Yu, and L.~Huang, ``Shape prior guided attack: Sparser perturbations on 3d point clouds,'' in \emph{Proceedings of the AAAI Conference on Artificial Intelligence}, vol.~36, no.~8, 2022, pp. 8277--8285.

\bibitem{dong2022isometric}
Y.~Dong, J.~Zhu, X.-S. Gao \emph{et~al.}, ``Isometric 3d adversarial examples in the physical world,'' \emph{Advances in Neural Information Processing Systems}, vol.~35, pp. 19\,716--19\,731, 2022.

\bibitem{brendel2017decision}
W.~Brendel, J.~Rauber, and M.~Bethge, ``Decision-based adversarial attacks: Reliable attacks against black-box machine learning models,'' \emph{arXiv preprint arXiv:1712.04248}, 2017.

\bibitem{brunner2019guessing}
T.~Brunner, F.~Diehl, M.~T. Le, and A.~Knoll, ``Guessing smart: Biased sampling for efficient black-box adversarial attacks,'' in \emph{Proceedings of the IEEE/CVF International Conference on Computer Vision}, 2019, pp. 4958--4966.

\bibitem{chen2020hopskipjumpattack}
J.~Chen, M.~I. Jordan, and M.~J. Wainwright, ``Hopskipjumpattack: A query-efficient decision-based attack,'' in \emph{2020 ieee symposium on security and privacy (sp)}.\hskip 1em plus 0.5em minus 0.4em\relax IEEE, 2020, pp. 1277--1294.

\bibitem{liu2022point}
D.~Liu, W.~Hu, and X.~Li, ``Point cloud attacks in graph spectral domain: When 3d geometry meets graph signal processing,'' \emph{arXiv preprint arXiv:2207.13326}, 2022.

\bibitem{maho2021surfree}
T.~Maho, T.~Furon, and E.~Le~Merrer, ``Surfree: a fast surrogate-free black-box attack,'' in \emph{Proceedings of the IEEE/CVF Conference on Computer Vision and Pattern Recognition}, 2021, pp. 10\,430--10\,439.

\bibitem{fan2017point}
H.~Fan, H.~Su, and L.~J. Guibas, ``A point set generation network for 3d object reconstruction from a single image,'' in \emph{Proceedings of the IEEE Conference on Computer Vision and Pattern Recognition (CVPR)}, 2017, pp. 605--613.

\bibitem{huttenlocher1993comparing}
D.~P. Huttenlocher, G.~A. Klanderman, and W.~J. Rucklidge, ``Comparing images using the hausdorff distance,'' \emph{IEEE Transactions on Pattern Analysis and Machine Intelligence}, vol.~15, no.~9, pp. 850--863, 1993.

\bibitem{yang2019pointflow}
G.~Yang, X.~Huang, Z.~Hao, M.-Y. Liu, S.~Belongie, and B.~Hariharan, ``Pointflow: 3d point cloud generation with continuous normalizing flows,'' in \emph{Proceedings of the IEEE/CVF international conference on computer vision}, 2019, pp. 4541--4550.

\bibitem{james1980monte}
F.~James, ``Monte carlo theory and practice,'' \emph{Reports on progress in Physics}, vol.~43, no.~9, p. 1145, 1980.

\bibitem{yang2019adversarial}
J.~Yang, Q.~Zhang, R.~Fang, B.~Ni, J.~Liu, and Q.~Tian, ``Adversarial attack and defense on point sets,'' \emph{arXiv preprint arXiv:1902.10899}, 2019.

\bibitem{hamdi2020advpc}
A.~Hamdi, S.~Rojas, A.~Thabet, and B.~Ghanem, ``Advpc: Transferable adversarial perturbations on 3d point clouds,'' in \emph{European Conference on Computer Vision (ECCV)}, 2020, pp. 241--257.

\bibitem{lou2024hide}
T.~Lou, X.~Jia, J.~Gu, L.~Liu, S.~Liang, B.~He, and X.~Cao, ``Hide in thicket: Generating imperceptible and rational adversarial perturbations on 3d point clouds,'' in \emph{Proceedings of the IEEE/CVF Conference on Computer Vision and Pattern Recognition}, 2024, pp. 24\,326--24\,335.

\bibitem{li2025nopain}
Z.~Li, X.~Du, N.~Lei, L.~Chen, and W.~Wang, ``Nopain: No-box point cloud attack via optimal transport singular boundary,'' in \emph{Proceedings of the Computer Vision and Pattern Recognition Conference}, 2025, pp. 3492--3502.

\bibitem{tang2026rethinking}
K.~Tang, X.~Liu, W.~Peng, X.~Wang, D.~Liu, P.~Zhu, C.~Lu, and Z.~Tian, ``Rethinking transferable adversarial attacks on point clouds from a compact subspace perspective,'' \emph{arXiv preprint arXiv:2601.23102}, 2026.

\bibitem{goyal2021revisiting}
A.~Goyal, H.~Law, B.~Liu, A.~Newell, and J.~Deng, ``Revisiting point cloud shape classification with a simple and effective baseline,'' in \emph{International Conference on Machine Learning}.\hskip 1em plus 0.5em minus 0.4em\relax PMLR, 2021, pp. 3809--3820.

\bibitem{xiang2021walk}
T.~Xiang, C.~Zhang, Y.~Song, J.~Yu, and W.~Cai, ``Walk in the cloud: Learning curves for point clouds shape analysis,'' in \emph{Proceedings of the IEEE/CVF International Conference on Computer Vision}, 2021, pp. 915--924.

\bibitem{ai2025upp}
Z.~Ai, Z.~Cui, Y.~Peng, and J.~Zhou, ``Upp: Unified point-level prompting for robust point cloud analysis,'' in \emph{Proceedings of the IEEE/CVF International Conference on Computer Vision}, 2025, pp. 27\,359--27\,368.

\bibitem{li2025learning}
Q.~Li, H.~Feng, X.~Gong, and Y.-S. Liu, ``Learning normals of noisy points by local gradient-aware surface filtering,'' in \emph{Proceedings of the IEEE/CVF International Conference on Computer Vision}, 2025, pp. 28\,828--28\,838.

\bibitem{li2022robust}
K.~Li, Z.~Zhang, C.~Zhong, and G.~Wang, ``Robust structured declarative classifiers for 3d point clouds: Defending adversarial attacks with implicit gradients,'' in \emph{Proceedings of the IEEE/CVF Conference on Computer Vision and Pattern Recognition}, 2022, pp. 15\,294--15\,304.

\bibitem{zhou2019dup}
H.~Zhou, K.~Chen, W.~Zhang, H.~Fang, W.~Zhou, and N.~Yu, ``Dup-net: Denoiser and upsampler network for 3d adversarial point clouds defense,'' in \emph{Proceedings of the IEEE International Conference on Computer Vision (ICCV)}, 2019, pp. 1961--1970.

\bibitem{googlestreetview}
\BIBentryALTinterwordspacing
{Google}, ``Google street view api,'' 2026. [Online]. Available: \url{https://google.com}
\BIBentrySTDinterwordspacing

\bibitem{spiess2024sketchfab}
F.~Spiess, R.~Waltensp{\~A}{\v{z}}l, and H.~Schuldt, ``The sketchfab 3d creative commons collection (s3d3c),'' \emph{arXiv preprint arXiv:2407.17205}, 2024.

\bibitem{sun2020scalability}
P.~Sun, H.~Kretzschmar, X.~Dotiwalla, A.~Chouard, V.~Patnaik, P.~Tsui, J.~Guo, Y.~Zhou, Y.~Chai, B.~Caine \emph{et~al.}, ``Scalability in perception for autonomous driving: Waymo open dataset,'' in \emph{Proceedings of the IEEE/CVF conference on computer vision and pattern recognition}, 2020, pp. 2446--2454.

\bibitem{chang2019argoverse}
M.-F. Chang, J.~Lambert, P.~Sangkloy, J.~Singh, S.~Bak, A.~Hartnett, D.~Wang, P.~Carr, S.~Lucey, D.~Ramanan \emph{et~al.}, ``Argoverse: 3d tracking and forecasting with rich maps,'' in \emph{Proceedings of the IEEE/CVF conference on computer vision and pattern recognition}, 2019, pp. 8748--8757.

\end{thebibliography}

\vspace{-10mm}
\begin{IEEEbiography}
[{\includegraphics[width=1in,height=1.25in,clip,keepaspectratio]{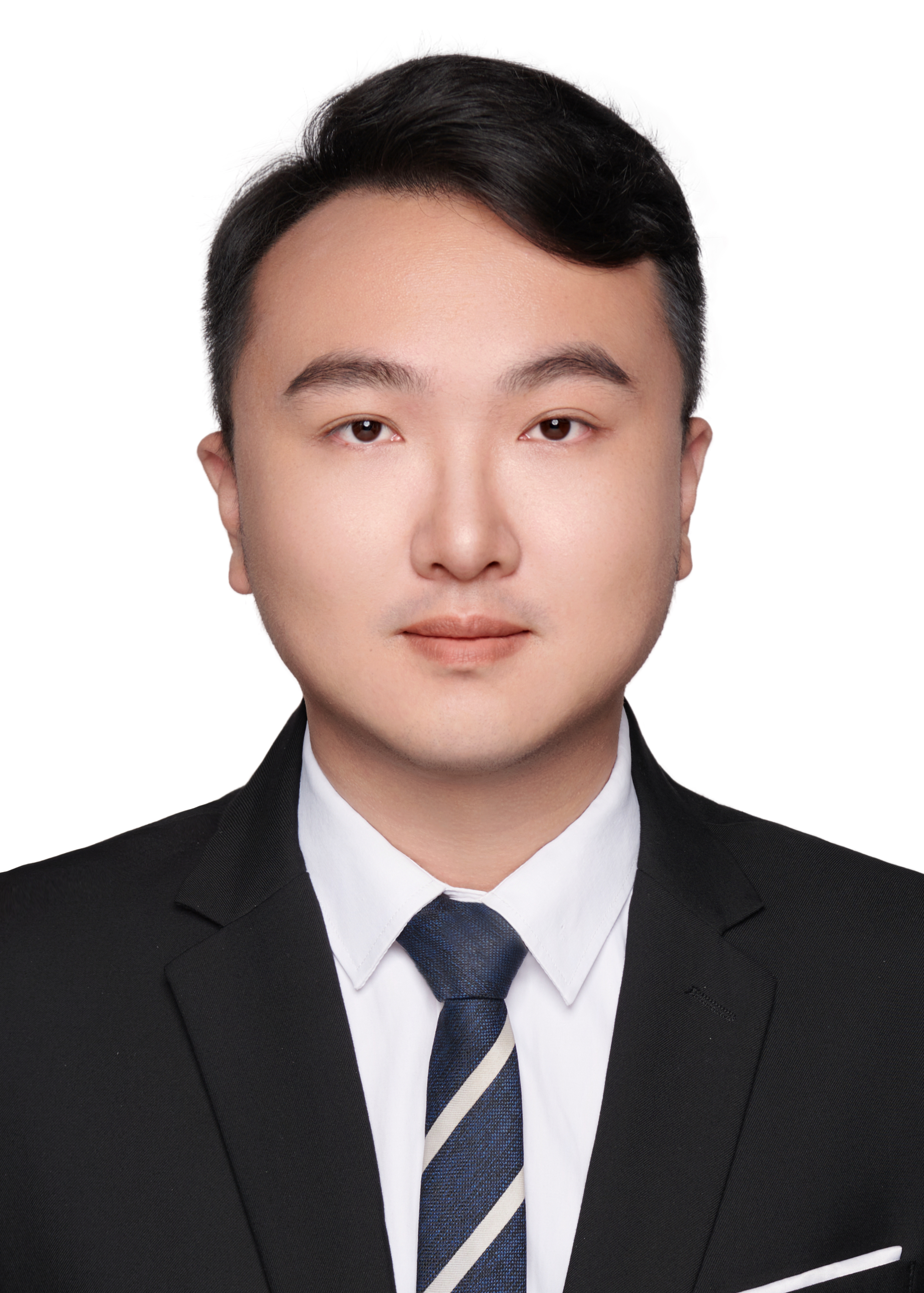}}]{Daizong Liu} is currently a tenure-track Assistant Professor in the Institute for Math \& AI at Wuhan University. He received the M.S. degree in Electronic Information and Communication of Huazhong University of Science and Technology in 2021, and the Ph.D. degree at Wangxuan Institute of Computer Technology of Peking University in 2025. His research interests include 3D adversarial attacks, multi-modal learning, LVLM robustness, etc. He has published more than 50 papers in refereed conference proceedings and journals such as TPAMI, TIFS, NeurIPS, ICLR, CVPR, ICCV, ECCV, SIGIR, AAAI, ACL. He regularly serves on the program committees of top-tier AI conferences such as NeurIPS, ICML, ICLR, CVPR, ICCV and ACL.
\end{IEEEbiography}

\vspace{-10mm}
\begin{IEEEbiography}
[{\includegraphics[width=1in,height=1.25in,clip,keepaspectratio]{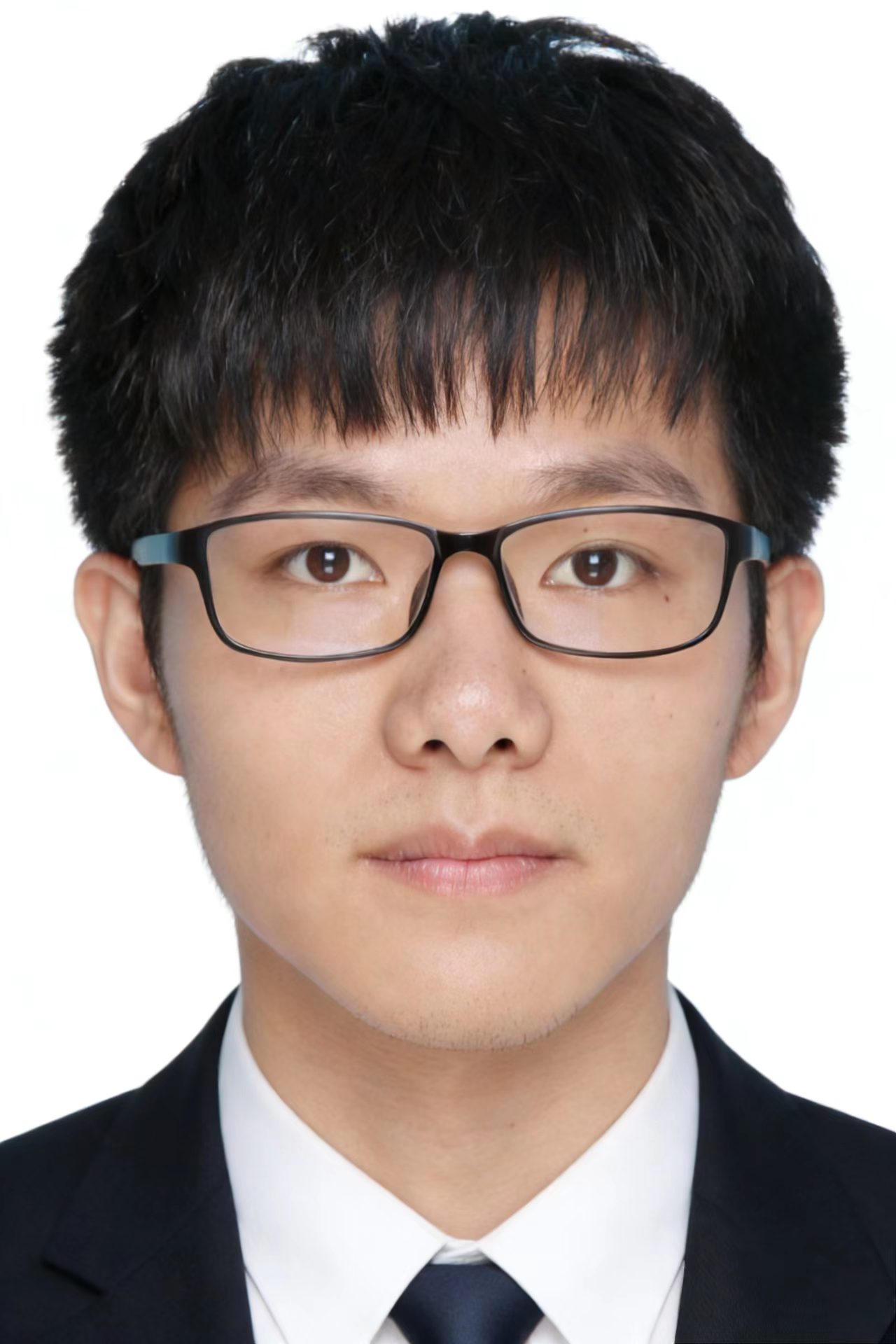}}]{Yunbo Tao} received his B.S. and M.S. degrees from the School of Cyber Science and Engineering, Huazhong University of Science and Technology. He is currently pursuing a Ph.D. degree at the College of Computer Science and Artificial Intelligence, Fudan University. His research interests include computer vision and AI security.
\end{IEEEbiography}

\vspace{-10mm}
\begin{IEEEbiography}[{\includegraphics[width=1in,height=1.25in,clip,keepaspectratio]{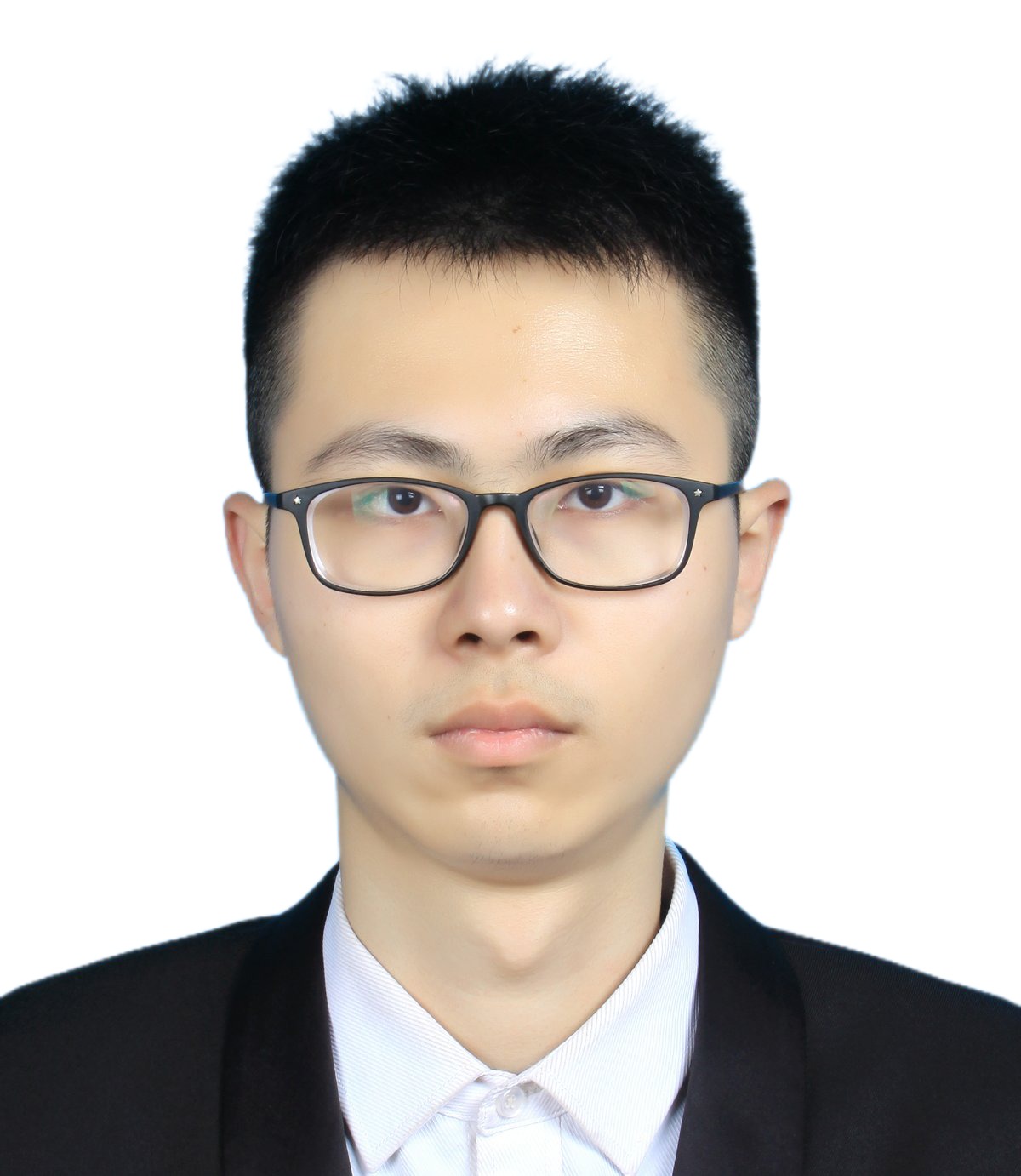}}]{Junhao Dong}
	received the M.S. degree in computer science and technology from Sun Yat-sen University, Guangzhou, China, in 2023. He is currently working toward the Ph.D. degree with the College of Computing and Data Science, Nanyang Technological University, Singapore. His research interests include trustworthy AI, computer vision, and adversarial machine learning. He is an assistant program chair of NeurIPS 2025.
\end{IEEEbiography}

\vspace{-10mm}
\begin{IEEEbiography}
[{\includegraphics[width=1in,height=1.25in,clip,keepaspectratio] {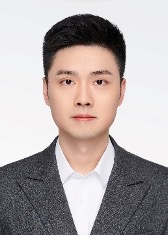}}]
 {Keke Tang} (Member, IEEE) received the B.Eng. degree from Jilin University, Changchun, China, in 2012, and the Ph.D. degree from the University of Science and Technology of China, Hefei, China, in 2017. He was a Post-Doctoral Fellow with The University of Hong Kong, Hong Kong. In 2019, he joined Guangzhou University, Guangzhou, China, where he is currently a Full Professor. His research interests include the areas of robotics, computer vision, computer graphics, and cyberspace security.
\end{IEEEbiography}

\vspace{-10mm}
\begin{IEEEbiography}[{\includegraphics[width=1in,height=1.25in,clip,keepaspectratio]{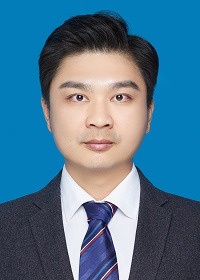}}]{Pan Zhou}
is currently a full professor and PhD
advisor with Hubei Engineering Research Center
on Big Data Security, School of Cyber Science
and Engineering, Huazhong University of
Science and Technology (HUST), Wuhan, P.R.
China. He received his Ph.D. in the School
of Electrical and Computer Engineering at the
Georgia Institute of Technology (Georgia Tech)
in 2011, Atlanta, USA. He received the ``Rising
Star in Science and Technology of HUST'' in
2017, and the ``Best Scientific Paper Award'' in
the 25th International Conference on Pattern Recognition (ICPR 2020).
He is currently an associate editor of IEEE Transactions on Network
Science and Engineering. His current research interest includes: security
and privacy, big data analytics, machine learning, and information
networks.
\end{IEEEbiography}

\vspace{-10mm}
\begin{IEEEbiography}
[{\includegraphics[width=1in,height=1.25in,clip,keepaspectratio] {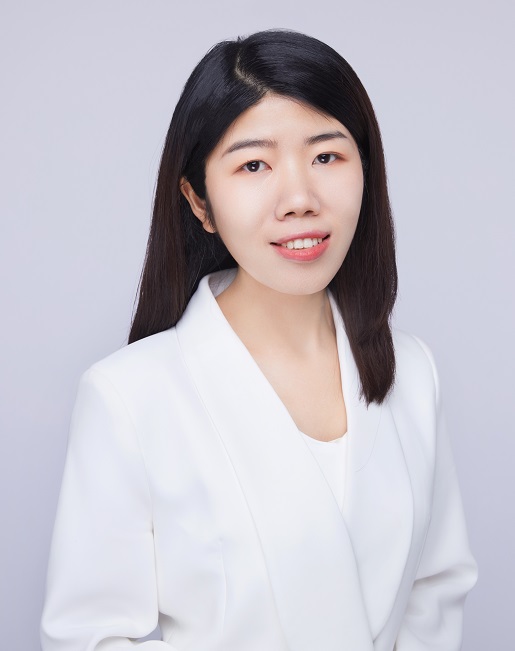}}]
  {Wei Hu}
  (M'17-SM'21) received the B.S. degree in Electrical Engineering from the University of Science and Technology of China in 2010, and the Ph.D. degree in Electronic and Computer Engineering from the Hong Kong University of Science and Technology in 2015.
 She was a Researcher with Technicolor, Rennes, France, from 2015 to 2017. She is currently an Associate Professor with Wangxuan Institute of Computer Technology, Peking University. Her research interests are graph signal processing, graph-based machine learning and 3D visual computing. She has authored around 90 international journal and conference publications, with several paper awards including Best Paper Candidate in CVPR 2021 and Best Student Paper Runner Up Award in ICME 2020. 
 She was awarded the 2021 IEEE Multimedia Rising Star Award---Honorable Mention, and the 2023 Outstanding Editorial Board Member Award in IEEE Signal Processing Society. She serves as an Associate Editor for IEEE Signal Processing Magazine and IEEE Transactions on Signal and Information Processing over Networks. She serves as a member of the Multimedia Signal Processing Technical Committee (MMSP-TC) and the Multimedia Systems \& Applications Technical Committee (MSA-TC).
 (\textit{Email: forhuwei@pku.edu.cn})
\end{IEEEbiography}

\vspace{-10mm}
\begin{IEEEbiography}[{\includegraphics[width=1in,height=1.25in,clip,keepaspectratio]{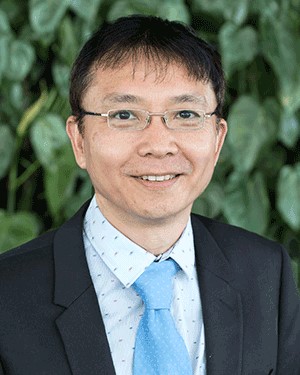}}]{Yew-Soon Ong}
	(Fellow, IEEE) received the Ph.D. degree in artificial intelligence in complex design from the University of Southampton, Southampton, U.K., in 2003. He is the President’s Chair Professor of Computer Science with Nanyang Technological University, Singapore, and the Chief Artificial Intelligence Scientist of the Agency for Science, Technology and Research, Singapore. He serves as the Co-Director of Singtel-NTU Cognitive and Artificial Intelligence Joint Lab. His research interest is in artificial and computational intelligence. Dr. Ong is the Founding Editor-in-Chief of the IEEE Transactions on Emerging Topics in Computational Intelligence and an Associate Editor of IEEE Transactions on Neural Networks and Learning Systems, IEEE Transactions on Cybernetics, and IEEE Transactions on Artificial Intelligence. He has received several IEEE outstanding paper awards and was listed as a Thomson Reuters Highly Cited Researcher and among the World’s Most Influential Scientific Minds.
\end{IEEEbiography}

\end{document}